\documentclass[10pt,twocolumn,letterpaper]{article}

\usepackage[pagenumbers]{cvpr} 

\definecolor{cvprblue}{rgb}{0.21,0.49,0.74}
\usepackage[pagebackref,breaklinks,colorlinks,allcolors=cvprblue]{hyperref}
\usepackage{graphicx}
\usepackage{multirow}
\usepackage{array}
\usepackage{booktabs}
\usepackage{tabularx}
\usepackage{comment}
\usepackage{colortbl}
\usepackage{bbm}
\usepackage{lscape}
\usepackage{here}
\usepackage{balance}

\def\paperID{12527} 
\def\confName{CVPR}
\def\confYear{2025}

\title{\textsc{ACT-Bench}: Towards Action Controllable World Models for \\ Autonomous Driving}

\author{
    Hidehisa Arai \quad Keishi Ishihara \quad Tsubasa Takahashi \quad Yu Yamaguchi \\
    Turing Inc.\\
    {\tt\small \{hidehisa.arai, keishi.ishihara\}@turing-motors.com}
}

\begin{document}

\twocolumn[{%
\renewcommand\twocolumn[1][]{#1}%
\maketitle
}]

\begin{abstract}

World models have emerged as promising neural simulators for autonomous driving, with the potential to supplement scarce real-world data and enable closed-loop evaluations. 
However, current research primarily evaluates these models based on visual realism or downstream task performance, with limited focus on fidelity to specific action instructions—a crucial property for generating targeted simulation scenes. 
Although some studies address action fidelity, their evaluations rely on closed-source mechanisms, limiting reproducibility.
To address this gap, we develop an open-access evaluation framework, \textsc{ACT-Bench}, for quantifying action fidelity, along with a baseline world model, \textsc{Terra}. 
Our benchmarking framework includes a large-scale dataset pairing short context videos from nuScenes with corresponding future trajectory data, which provides conditional input for generating future video frames and enables evaluation of action fidelity for executed motions. 
Furthermore, \textsc{Terra} is trained on multiple large-scale trajectory-annotated datasets to enhance action fidelity.
Leveraging this framework, we demonstrate that the state-of-the-art model does not fully adhere to given instructions, while \textsc{Terra} achieves improved action fidelity.
\textbf{All components of our benchmark framework will be made publicly available to support future research.}
\end{abstract}

\section{Introduction}
\label{sec:intro}

\begin{figure}[htbp]
    \centering
    \includegraphics[width=\linewidth]{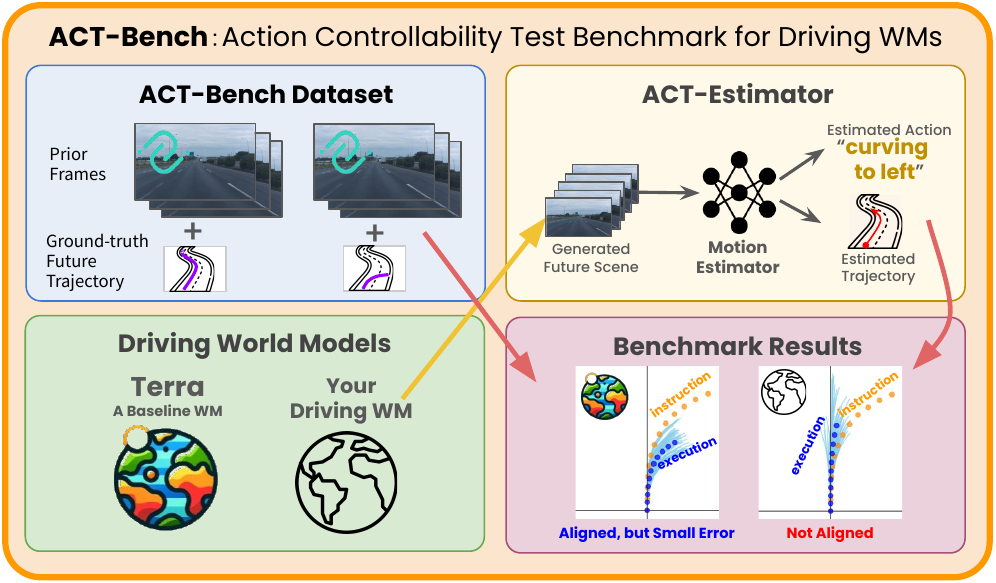}
    \caption{\textbf{\textsc{ACT-Bench}}: Action Controllability Test Benchmark. This benchmark suite evaluates action fidelity in driving world models by utilizing a unique dataset consisting of prior frames paired with ground-truth future trajectories. It enables systematic assessment of action fidelity and comparison with the novel world model, \textsc{Terra}.}
    \label{fig:fail-example}
\end{figure}

\begin{figure*}[htbp]
    \centering
    \includegraphics[width=\textwidth]{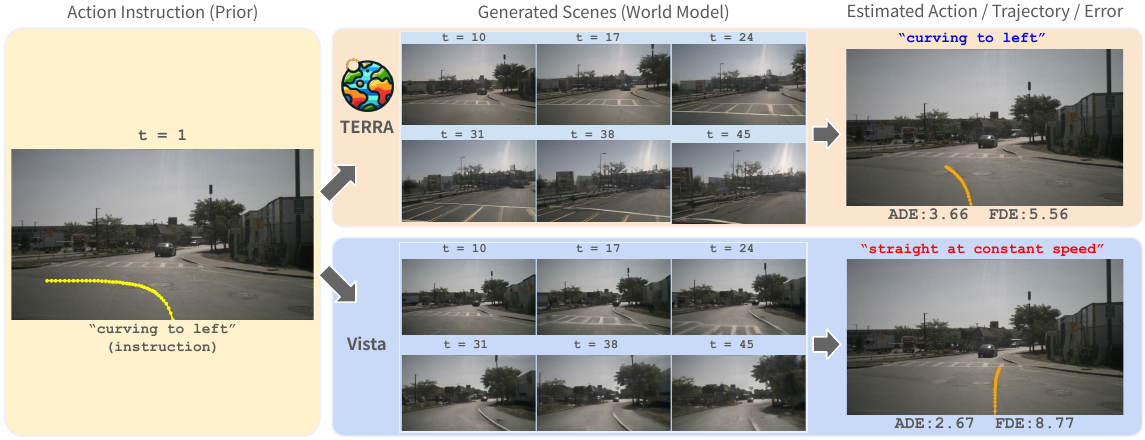}
    \caption{\textsc{ACT-Bench} assesses the action controllability of world models by estimating actions, trajectories, and their deviations from the generated driving scenes using our motion estimator, \textsc{ACT-Estimator}. In the upper example, \textsc{Terra} successfully follows the instruction to ``curving to left." In contrast, the lower example illustrates that Vista fails to follow the instruction. This evaluation helps to identify cases where driving world models do not adhere to the given action instructions, and compare performance for different models.}
    \label{fig:fail-example}
\end{figure*}

The field of autonomous driving has witnessed rapid advancements in recent years, driven by progress in perception, predictive capabilities, and decision-making~\cite{yurtsever2020survey,grigorescu2020survey,chen2024end}. These developments aim to realize a future where autonomous vehicles can safely navigate complex urban environments and respond reliably to dynamic conditions. However, achieving this goal requires overcoming significant challenges, such as gathering extensive real-world data that includes safety-critical scenarios, or continuously adapting to unpredictable road conditions.

One promising approach to addressing these challenges is through the use of world models~\cite{ha2018worldmodels, lecun2022path}. 
In autonomous driving, world models serve as neural simulators~\cite{zhu2024sora} that generate synthetic environments and scenarios, supplementing real-world data, which can be difficult or dangerous to collect, and enabling closed-loop evaluations of autonomous driving systems. 
Early research explored the use of world models on simulator as policy models~\cite{pan2022iso,hu2022model,gao2024enhance}, directly guiding decision-making. 
However, due to the excessive complexity of the modeling target, the focus has shifted toward employing world models as simulators to generate synthetic data~\cite{wang2023drivedreamer,hu2023gaia,wang2024driving,lu2025wovogen} and evaluate policy models~\cite{wang2024driving,gao2024vista}.

Despite these advancements, practical applications still lack fidelity to action-conditioned instructions—a crucial factor for reliable simulations, especially in safety-critical applications. Recently, 
Vista~\cite{gao2024vista} introduced an action fidelity-aware model that allows for conditional inputs specifying a wide range of actions.
Although Vista accounts for action fidelity, it generates scenes fail to fully adhere to the given instructions (Figure \ref{fig:fail-example}). 
This failure suggests that its fidelity performance remains insufficient.

To further advance action fidelity-aware world models, open-access evaluation benchmarks are needed. 
GenAD~\cite{yang2024genad} is the first work to measure fidelity by comparing estimated and ground truth trajectories in generated scenes.
However, this metric relies on a non-public evaluation model, which limits reproducibility. 
Although Vista~\cite{gao2024vista} has made its world model public, it still depends on a closed benchmark. 
In addition to evaluation metrics, there are very few open-access baseline models. While GAIA-1~\cite{hu2023gaia} is a well-known world model, only a technical report has been published, without access to substantial resources or models. 
This lack of open-access benchmarks and baseline models restricts broader research efforts in this domain. 
Addressing these gaps by creating accessible benchmarks and models would not only foster innovation but also support standardized evaluations and promote reproducible studies in driving world models.

In response to this gap, we first present a benchmark suite, \textsc{ACT-Bench} ({\bf A}ction {\bf C}ontrollability {\bf T}est {\bf Bench}mark), to evaluate action fidelity in driving world models. 
To establish this benchmark, we construct a unique dataset by annotating high-level actions, such as ``curving to right," and their corresponding trajectories for short driving scenes in the widely-used nuScenes~\cite{caesar2020nuscenes} dataset.
These annotated actions serve as ground truth for each short driving scene in nuScenes, enabling systematic assessment of action fidelity in driving world models.
To estimate trajectories from the generated driving scenes, we develop a motion estimator, \textsc{ACT-Estimator}, to systematically evaluate how accurately a world model follows specific driving instructions. 
The estimator quantifies both the consistency between the intended action (i.e., instruction) and its execution, as well as the precision and accuracy of the action trajectory in spatial coordinates by comparing it to the ground truth.
The estimator model is a key contribution, enabling automatic fidelity assessment of driving world models with respect to action instructions.

Second, we introduce a baseline world model, \textsc{Terra}, designed for flexible control based on target trajectories. 
The model architecture of \textsc{Terra} follows GAIA-1, but as GAIA-1 is not publicly accessible, \textsc{Terra} represents the first open-access model sharing the GAIA-1's philosophy. 
To enhance action fidelity, \textsc{Terra} is trained on three datasets: OpenDV-YouTube~\cite{yang2024genad}, nuScenes~\cite{caesar2020nuscenes} and CoVLA~\cite{arai2024covla}, a large-scale dataset annotated with corresponding future trajectories. 
This expanded dataset allows \textsc{Terra} to better capture action fidelity, leveraging more annotated scenes than Vista, which was trained on two datasets, OpenDV-YouTube and nuScenes.

Using the proposed benchmark and baseline model, we examine how well existing world models follow instructions. 
First, we confirm that our evaluator model achieves sufficient performance in assessment tasks.
Our empirical studies show that \textsc{Terra} outperforms Vista, although both models demonstrate adequate quality across various metrics. 
Notably, in terms of the match rate between instructed high-level actions and their executions, our experimental study confirms that Vista achieves a 30.72\% match, while \textsc{Terra} reaches 44.11\%.
This finding indicates that achieving high fidelity in driving world models remains challenging, and significant improvements will require collaborative efforts and broader discussions within the research community.

\begin{table*}[t]
\centering
\footnotesize
\begin{tabular}{l|cccc}
\hline
\textbf{Method} & \multicolumn{4}{c}{\textbf{Evaluation Metrics}} \\
 & \textbf{Visual Quality} & \textbf{Layout Controllability} & \textbf{Downstream Task} & \textbf{Action Fidelity} \\ \hline
\cellcolor[gray]{0.9}GAIA-1~\cite{hu2023gaia} & N/A & - & N/A & N/A \\ 
\cellcolor[gray]{0.9}DriveDreamer~\cite{wang2023drivedreamer} & FID, FVD & N/A & 3D object detection, planning & N/A \\ 
\cellcolor[gray]{0.9}WoVoGen~\cite{lu2025wovogen} & FID, FVD & N/A & N/A & N/A \\
\cellcolor[gray]{0.9}ADriver-I~\cite{jia2023adriver} & FID, FVD & - & planning & N/A \\ 
\cellcolor[gray]{0.9}DriveWM~\cite{wang2024driving} & FID, FVD & Map\&Object & 3D object detection, planning & N/A \\ 
\cellcolor[gray]{0.9}GenAD~\cite{yang2024genad} & FID, FVD, CLIPSIM & - & planning & \cellcolor[gray]{0.9}VO-based \\ 
Vista~\cite{gao2024vista} & FID, FVD & - & N/A & \cellcolor[gray]{0.9}VO-based \\ \hline

\end{tabular}
\caption{\textbf{Comparison of Evaluation Metrics used for Autonomous Driving World Models.} ``N/A" indicates that either the evaluation has not been conducted or that only qualitative assessment is available, while ``-" signifies that evaluation could not be conducted due to input limitations of the respective model. \colorbox[gray]{0.9}{Highlighted models and evaluation metrics} are not publicly available.}
\label{tab:world_model_comparison}
\end{table*}

Our contributions are summarized in three-folds:
\begin{itemize}
    \item We present a novel benchmark, \textsc{ACT-Bench}, that evaluates action fidelity in driving world models by constructing a unique dataset based on the nuScenes dataset. This dataset, annotated with high-level driving actions and trajectories, enables systematic assessment of a model's ability to execute specific driving instructions.
    \item We introduce \textsc{Terra}, a driving world model designed for flexible trajectory control, based on GAIA-1 architecture but trained on large scale trajectory annotated datasets.
    \item Using our benchmark suite, we demonstrate that the state-of-the-art model, Vista, does not fully adhere to given instructions, while \textsc{Terra} demonstrates improved action fidelity. 
\end{itemize}

\section{Related Work}\label{sec:relatedwork}

This section reviews the existing driving world models and evaluation metrics they utilized.
The brief summary of these world models and metrics are summarized in Table~\ref{tab:world_model_comparison}.

\subsection{Driving World Model}
World models offer agents a latent representation of the environment, enabling them to simulate potential futures and explore outcomes of various actions within this learned space~\cite{ha2018worldmodels, hu2023toward, lecun2022path, zhu2024sora}. Rather than directly coupling perception and action, world models create an abstract internal environment that agents can use to forecast and evaluate sequences of actions. This predictive capability allows models to simulate and evaluate complex scenarios efficiently, in domains like representation learning~\cite{wu2024pre,garrido2024learning,bardes2024vjepa,gupta2022maskvit,schwarzer2020data,wu2023masked}, model-based reinforcement learning~\cite{ha2018worldmodels,hafner2019dream,hafner2020mastering,hafner2023mastering,seo2023masked,wu2023daydreamer,sekar2020planning}, and model predictive control~\cite{finn2017deep,hafner2019learning,mendonca2023structured,huang2024safedreamer,ebert2018visual}, supporting advancements in areas such as game AI~\cite{bruce2024genie,Kaiser2020Model,hafner2020mastering,hafner2023mastering,micheli2023transformers,robine2023transformerbased,zhang2024storm,alonso2024diffusion} and robotics~\cite{hafner2019learning,hafner2019dream,huang2024safedreamer,wu2023daydreamer,piergiovanni2019learning,mendonca2023structured,ma2024harmonydream,yang2024learning}.

Over the past year, world models for autonomous driving have emerged and rapidly evolved, with recent advancements introducing models that emphasize different strengths: high-quality video generation~\cite{hu2023gaia,gao2024vista,jia2023adriver,yang2024genad}, consistent multi-view outputs~\cite{wang2023drivedreamer,wang2024driving,lu2025wovogen}, and the ability to incorporate diverse conditional inputs~\cite{hu2023gaia,yang2024genad,lu2025wovogen,wang2023drivedreamer,wang2024driving,jia2023adriver}, such as text prompts, bounding boxes, and map information. Together, these developments enable visually realistic driving simulations across various scenarios, significantly expanding their utility for training and testing autonomous systems in more flexible and robust ways.

\subsection{Evaluation Metrics for Driving World Model}

{\bf Visual Quality Metric.}
A core task of driving world model is video generation task.
In recent years, evaluation of video generation models has commonly relied on metrics such as Fréchet Inception Distance (FID)~\cite{heusel2017gans}, Fréchet Video Distance (FVD)~\cite{unterthiner2018towards,unterthiner2019fvd}, and CLIPSIM~\cite{radford2021learning}. 

\noindent
{\bf Action Fidelity.}
Evaluation of how faithfully generated content adheres to action-based conditioning was first conducted by GenAD~\cite{yang2024genad}. This approach utilizes an XVO~\cite{lai2023xvo}-based Monocular Visual Odometry~\cite{nister2006visual} (VO) algorithm to estimate the given instruction from the generated video, then compares the estimated instruction with the actual input to assess fidelity. A similar method is employed in Vista~\cite{gao2024vista}; however, the VO algorithm used for instruction estimation and the evaluation dataset are not publicly available, meaning open evaluation has not yet been possible.

\noindent
{\bf Other Metrics.}
DriveDreamer~\cite{wang2023drivedreamer} and DriveWM~\cite{wang2024driving} use layout information such as bounding boxes or HD-Maps for conditioning, often assessing fidelity to the instructed layout through 3D object detection. 
Additionally, these world models are sometimes used to predict future trajectories for planning purposes or to augment data for perception tasks like 3D object detection. 
In such cases, evaluation often focuses on performance in downstream tasks.

\section{ACT-Bench}
\label{sec:action_fidelity_evaluation}

Towards establishing an open benchmark for driving world models, particularly in terms of action fidelity for generated driving scenes, we present \textbf{A}ction \textbf{C}ontrollability \textbf{T}est \textbf{Bench}mark (\textsc{ACT-Bench}), which includes a driving scene dataset annotated with high-level actions and a mechanism for evaluating fidelity. Additionally, \textsc{ACT-Bench} incorporates a baseline world model \textsc{Terra}, providing a reference for evaluating the performance of other models.

In the following, we explain the benchmark dataset that captures diverse driving scenarios (Sec.~\ref{subsec:benchmark_dataset}), introduce benchmark metrics as indicators for evaluating action fidelity (Sec.~\ref{subsec:benchmark_metrics}), describe an automated system for objectively scoring the alignment of generated scenes with specified instructions (Sec.~\ref{subsec:automated_evaluation}), and present the baseline world model \textsc{Terra} to provide a reference point for evaluating action fidelity in driving scene generation (Sec.~\ref{subsec:terra}).

\subsection{Benchmark Dataset}
\label{subsec:benchmark_dataset}

We construct a benchmark dataset designed specifically to assess how well generated frame sequences align with given trajectory instructions. This dataset leverages a subset of the validation split from the widely used nuScenes~\cite{caesar2020nuscenes} dataset, augmented with trajectory templates for precise conditional generation tasks.

Our dataset comprises short video segments captured from the \texttt{CAM\_FRONT} camera in nuScenes. Although nuScenes contains various data modalities, such as multi-camera footage and LiDAR point clouds, we limit our focus to the front-facing camera view, as it captures the vehicle's immediate forward path -- essential for action-following evaluation and sufficient as a baseline for our purposes. Each video segment is paired with one or more trajectory templates, allowing a single  context video to support multiple trajectory conditions.
To obtain relevant video segments, we extract short intervals from 20-second nuScenes scenes, focusing on sequences where specific trajectory templates can be applied.

The trajectory templates are divided into nine categories. Each category covers a distinct driving maneuver, ensuring that the templates provide a broad spectrum of actions. Details of each category and the corresponding number of video-trajectory pairs are listed in Table~\ref{tab:action_counts}. Within each category, we provide options with varying degrees of curvature and speed, ensuring that the templates cover a broad spectrum of driving maneuvers. Each extracted interval consists of a 10-frame context video segment paired with these trajectory instructions, forming a basis for action-conditioned simulation evaluation.

To refine the dataset, we apply two levels of filtering. First, we remove video segment - trajectory pairs where the initial speed in the context video deviated from the template's starting speed by more than 10 km/h. Second, through visual inspection, we filter out cases where interactions with objects in front could affect evaluation or generation. This process results in a benchmark dataset of 2,286 video-trajectory pairs, each carefully selected to support consistent evaluation.

Furthermore, to accommodate world models that operate on a per-frame action-input basis - where actions are input at each simulation cycle to mimic autonomous planning models - we ensure that the trajectory templates were prepared for each frame in the sequence. In doing so, we adjusted each template to account for the ideal vehicle orientation at each step, aligning with the action's natural progression. To achieve the multi-frame trajectories, we extract trajectory data from the CoVLA dataset~\cite{arai2024covla}, which includes a diverse range of non-linear driving actions, allowing us to construct trajectories that encompass varied behaviors beyond simple linear motions.

\subsection{Benchmark Metrics}
\label{subsec:benchmark_metrics}

To capture different aspects of action fidelity, we introduce two distinct metrics: \textit{instruction-execution consistency} (IEC) and \textit{trajectory alignment} (TA).
IEC quantifies the degree of alignment between the given instructions and the executed actions, while TA measures the distance between the estimated trajectory and its ground truth corresponding to the given instruction.

\begin{table}[t]
    \centering
    \caption{\#Video-trajectory pairs for each action category.}
    \label{tab:action_counts}
    \footnotesize
    \begin{tabular}{lr}
        \toprule
        \textbf{High-level Action Category} & \textbf{Number of Pairs} \\
        \midrule
        \texttt{curving to left} & 162 \\
        \texttt{curving to right} & 188 \\
        \texttt{shifting towards left} & 115 \\
        \texttt{shifting towards right} & 192 \\
        \texttt{starting} & 89 \\
        \texttt{stopping} & 508 \\
        \texttt{accelerating} & 273 \\
        \texttt{straight at constant speed} & 541 \\
        \texttt{decelerating} & 218 \\
        \midrule
        \textbf{Total} & \textbf{2286} \\
        \bottomrule
    \end{tabular}
\end{table}

\noindent {\bf Instruction-Execution Consistency.}
We define the metric \textit{Instruction-Execution Consistency (IEC)}, which quantifies the degree of alignment between the given instructions and the executed actions.
Let $a^{\text{ins}}$ represent the instruction provided as a prior for generating scenes from the world model, and $a^{\text{est}}$ represent the estimated action derived from the generated scenes. Both $a^{\text{ins}}$ and $a^{\text{est}}$ belong to the set of high-level actions, denoted as $\mathcal{A}$, such that $a^{\text{ins}}, a^{\text{est}} \in \mathcal{A}$.
For $n$ samples, IEC can be assessed as follows:
\begin{equation}
\text{IEC} = \frac{1}{n} \sum_{j=1}^{n} \mathbbm{1}\{a^{\text{ins}}_j = a^{\text{est}}_j\}
\end{equation}
where $\mathbbm{1}\{ \cdot \}$ is the indicator function, which returns 1 if $a^{\text{ins}}_j$ and $a^{\text{est}}_j$ match, and 0 otherwise.

\noindent {\bf Trajectory Alignment.}
We also introduce the metric \textit{Trajectory Alignment (TA)}, which captures the precision and quality of action behavior, based on the instruction or objective.
TA measures how close the intended trajectory is to the estimated trajectory.
Let $\tau^{\text{ins}} \in\mathbb{R}^{T\times d}$ denote the intended trajectory provided as a conditioning signal, where $T$ represents the number of points in the trajectory and $d$ represents the dimensionality of each point. 
Similarly, let $\tau^{\text{est}} \in \mathbb{R}^{T \times d}$ denote the estimated trajectory derived from the generated video. 
The alignment between $\tau^{\text{ins}}$ and $\tau^{\text{est}}$ is quantified using a distance function $D: (\mathbb{R}^{T \times d}, \mathbb{R}^{T \times d}) \rightarrow \mathbb{R}$ that takes two trajectories as input and returns a scalar value, expressed as  
$D(\tau^{\text{ins}}, \tau^{\text{est}})$.
Common choices for $D$ include Average Displacement Error (ADE) and Final Displacement Error (FDE)~\cite{phong2023what}, which provide meaningful evaluations of trajectory closeness.
\begin{equation}
\label{eq:ade}
\text{ADE} = \frac{1}{T} \sum_{t=1}^{T} \|\tau^{\text{ins}}_t - \tau^{\text{est}}_t\|_2
\end{equation}
\begin{equation}
\label{eq:fde}
\text{FDE} = \|\tau^{\text{ins}}_T - \tau^{\text{est}}_T\|_2
\end{equation}
A smaller value indicates closer alignment.

\subsection{ACT-Estimator}\label{subsec:automated_evaluation}
Using the benchmarking dataset introduced above, we design a method to evaluate the action fidelity of world models.
Our evaluation relies on a model that performs both video classification to identify high-level action commands and vehicle trajectory estimation based on a sequence of generated camera frames.
This model serves as the basis for our automated evaluation metric.

\noindent{\bf Dataset.} The dataset used to train the \textsc{ACT-Estimator} model is constructed independently of the \textsc{ACT-Bench} benchmark dataset, with some differences in action classes. Specifically, due to the limited occurrence of actions like shifting towards left or right, these classes are excluded from the target classes of \textsc{ACT-Estimator}. Additionally, constant-speed scenes, which make up most of the nuScenes dataset, are divided into two classes based on speed. Each sample comprises a four-second sequence of frames from the nuScenes dataset, along with corresponding trajectory data and high-level action class labels. The trajectory data is derived from \texttt{ego\_pose} information from the \texttt{CAM\_FRONT} sensor, and action labels are generated using a rule-based algorithm categorizing trajectories into nine classes. Details on the labeling methodology are in the Appendix.

\noindent {\bf Joint Optimization.}
The \textsc{ACT-Estimator} is optimized as a multi-task framework, jointly learning to classify high-level vehicle movements (such as curving to left or right) and predict vehicle trajectories. This approach aims to improve classification accuracy for high-level actions by leveraging shared representations learned from trajectory estimation.
Each task has its respective loss function, which is combined to produce the final loss for training.

\begin{equation}
    \mathcal{L}_{\mathrm{total}} = \beta\cdot\mathcal{L}_{\mathrm{classification}} + (1 - \beta)\cdot\mathcal{L}_{\mathrm{trajectory}}
\end{equation}
where $\beta$ is a weighting factor that controls the relative importance of each task. 
We employ the cross-entropy loss as $\mathcal{L}_\mathrm{classification}$ and smooth L1 loss~\cite{girshick2015fast} as $\mathcal{L}_\mathrm{trajectory}$.

\noindent {\bf Model Architecture.}
The \textsc{ACT-Estimator} utilizes the proven I3D~\cite{carreira2017quo} architecture as its CNN backbone, extracting spatiotemporal features from input videos. 
These features are then flattened and passed through self-attention~\cite{vaswani2017attention} layers to focus on critical parts within the video. 
Following this, the processed information is passed to dual-task heads, each tailored for one of the two tasks.
\begin{enumerate}
     \item Classification Head: A multi-layer perceptron (MLP) predicts high-level action classes based on the features pooled through Global Average Pooling~\cite{lin2013network}.
     \item Trajectory Prediction Head: A GRU~\cite{cho-etal-2014-learning}-based unit with cross-attention~\cite{DBLP:journals/corr/BahdanauCB14} that autoregressively predicts 2D trajectory coordinates $(x, y)$ for each point in the path.
\end{enumerate}
The architecture of \textsc{ACT-Estimator} is shown in Figure \ref{fig:evaluator_model_architecture} (in Appendix).

\subsection{\textsc{Terra}: A Baseline World Model}
\label{subsec:terra}
As a baseline world model, we introduce \textsc{Terra}, an open-access world model designed for flexible trajectory control. \textsc{Terra} shares the same design principles as GAIA-1~\cite{hu2023gaia}, yet it is trained on three open datasets: OpenDV-YouTube~\cite{yang2024genad}, nuScenes~\cite{caesar2020nuscenes} and CoVLA datasets~\cite{arai2024covla}. Notably, \textsc{Terra} allows trajectory-based instructions to be input at each frame during conditioning, enabling precise control over the generated video. The brief architecture of \textsc{Terra} is illustrated in Figure \ref{fig:architecture_overview}, and additional details are provided in the Appendix.

\begin{figure}[t]
    \centering
    \includegraphics[width=.48\textwidth]{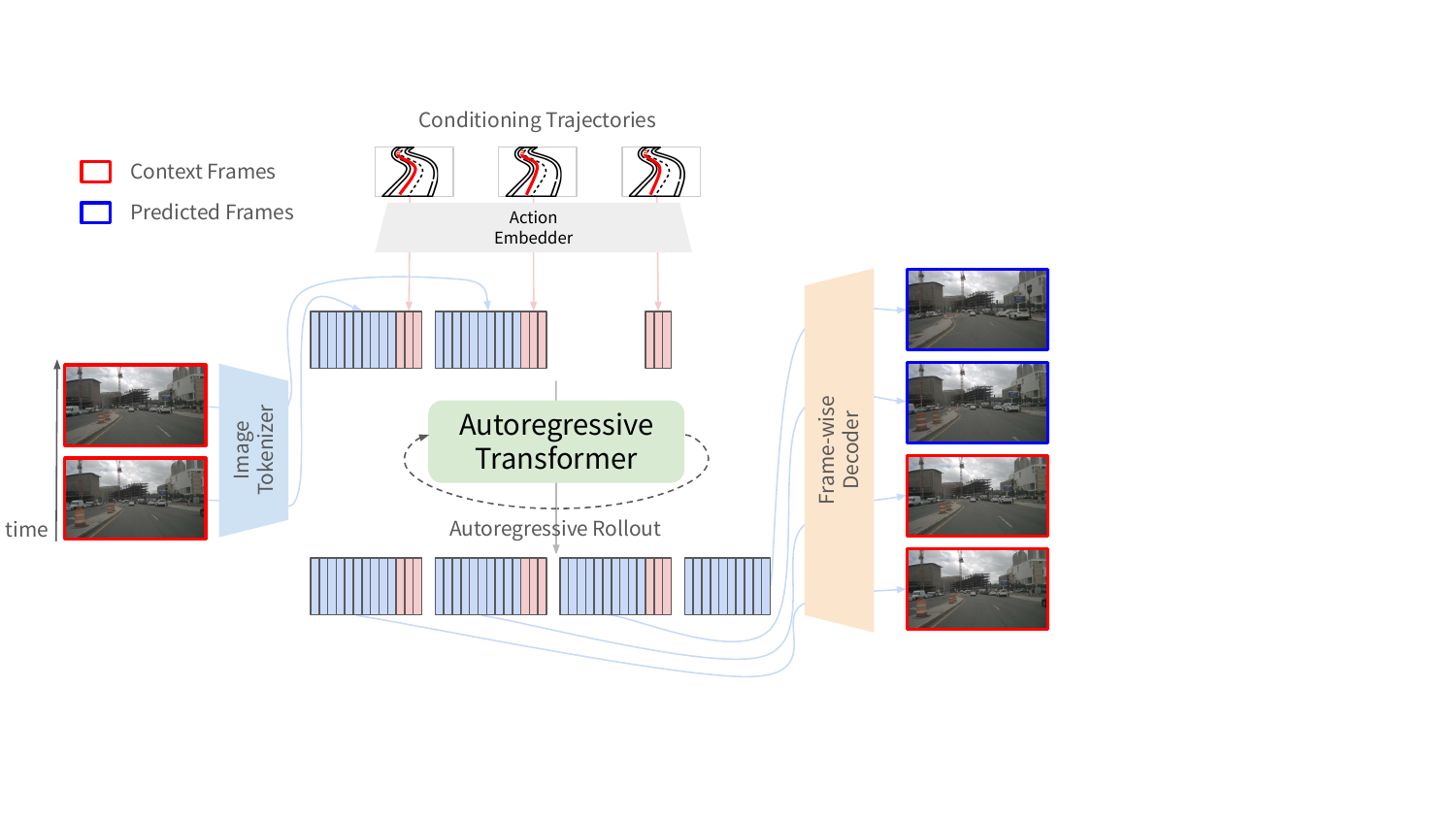}
    \caption{\textbf{\textsc{Terra}'s architecture overview.} \textsc{Terra} follows the same design philosophy as GAIA-1~\cite{hu2023gaia} but omits text conditioning capability and the use of video decoder to maintain simplicity.}
    \label{fig:architecture_overview}
\end{figure}
\section{Experiment: Estimator Validation}
\label{sec:eval_val}

Here, we validate the performance of \textsc{ACT-Estimator} by assessing how reliably it performs against the ground truth and related evaluation methods.

\noindent {\bf Training procedure.} The \textsc{ACT-Estimator} is trained for 30,850 iterations on four H100 GPUs with a per-GPU batch size of 12 and gradient accumulation steps of 2, resulting in an effective batch size of 96. We use the AdamW optimizer \cite{loshchilov2017decoupled} along with a OneCycleLR scheduler, setting the maximum learning rate to $1.2 \times 10^{-4}$.

\noindent {\bf Validation dataset.}
The training data for the \textsc{ACT-Estimator} follows the dataset described in Section \ref{subsec:automated_evaluation}, and the evaluations in the following subsections are conducted on a validation split created by randomly selecting 8,407 samples from this dataset. This validation split is solely used for evaluation purposes and is not included in the training of the \textsc{ACT-Estimator}.

\subsection{High-level Action Classification}

\begin{figure}[t]
    \centering
    \includegraphics[width=.85\linewidth]{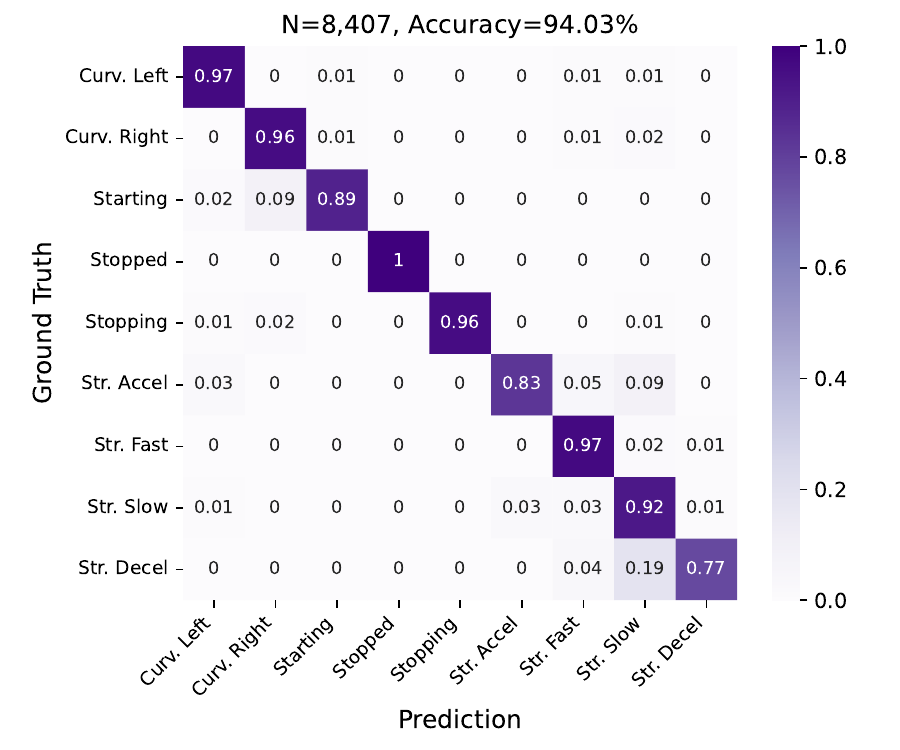}
    \caption{\textbf{Prediction accuracy of the \textsc{ACT-Estimator}} is visualized in this matrix across high-level actions on the validation dataset. Diagonal values indicate correct predictions, while off-diagonal represent mismatches. The overall accuracy is 94.03\%.}
    \label{fig:act_estimator_classification_performance}
\end{figure}

Figure \ref{fig:act_estimator_classification_performance} illustrates a high-level action classification task result. 
The classification results exhibit an accuracy exceeding \textbf{94\%}, indicating a strong capability in accurately identifying the intended actions. Furthermore, the \textsc{ACT-Estimator} demonstrates consistently high performance across all classes, showing robustness in distinguishing between various driving maneuvers. This consistent performance across classes underscores the model's reliability in capturing the nuances of different driving actions.

\begin{figure}[t]
    \centering
    \includegraphics[width=1.\linewidth]{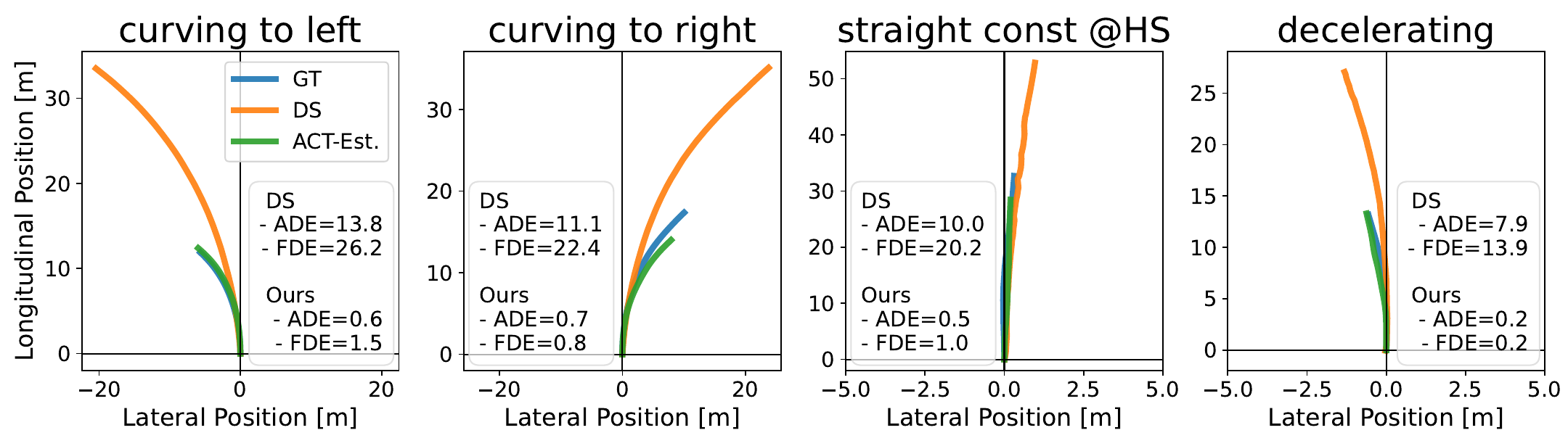}
    \caption{\textbf{Examples of Estimated Vehicle Trajectories.} DROID-SLAM (DS) tends to overestimate the trajectory length, especially in curving and straight at high-speed scenarios. Our model demonstrates higher alignment with the ground truth (GT) trajectory.} 
    \label{fig:traj_examples}
\end{figure}

\subsection{Vehicle Trajectory Estimation}

We here measure ADE (\ref{eq:ade}) and FDE (\ref{eq:fde}) to quantify how the estimated trajectories align against their ground truth. As a baseline for comparison, we employ DROID-SLAM~\cite{teed2021droid}, a well-established SLAM~\cite{durrant2006simultaneous} approach. To adapt DROID-SLAM to our monocular setup, we combine it with Metric3D~\cite{hu2024metric3d,yin2023metric3d} for monocular depth estimation.

Vista~\cite{gao2024vista} and GenAD~\cite{yang2024genad} utilize an XVO~\cite{lai2023xvo}-based mechanism, namely inverse dynamics estimation, to infer the trajectory from the generated scenes; however, the details of the mechanism are not disclosed yet (as discussed in Table \ref{tab:world_model_comparison}). Due to this, we employ DROID-SLAM as the baseline even though there is a target domain gap.

Table \ref{tab:ade_fde_comparison} show that our model outperforms this adapted DROID-SLAM setup on both ADE and FDE metrics, indicating superior accuracy in predicting vehicle trajectories. 
This result highlights \textsc{ACT-Estimator}'s capability to provide precise trajectory assessments, ensuring that the model is well-suited for trajectory alignment evaluation.

Figure \ref{fig:traj_examples} illustrates four estimated trajectories. DROID-SLAM tends to overestimate the trajectory length, especially in curving and straight high-speed scenarios. Our model demonstrates higher alignment with the ground truth trajectory, as reflected by lower ADE and FDE values.

\begin{table}[t]
\centering
\footnotesize
\begin{tabular}{l|cc|cc}
\toprule
\multirow{2}{*}{High-level Action Category} & \multicolumn{2}{c|}{ADE($\downarrow$)} & \multicolumn{2}{c}{FDE($\downarrow$)} \\
                       & Ours & DS & Ours & DS \\ \midrule
\texttt{curving to left}              & \textbf{0.82}            & 9.91            & \textbf{1.61}            & 18.70          \\
\texttt{curving to right}             & \textbf{0.78}            & 9.51            & \textbf{1.54}            & 18.44          \\
\texttt{starting}                     & \textbf{0.81}            & 3.25            & \textbf{1.61}            & 8.85           \\
\texttt{stopped}                      & 0.76                     & \textbf{0.06}   & 1.48                     & \textbf{0.09}  \\
\texttt{stopping}                     & \textbf{0.73}            & 7.15            & \textbf{1.41}            & 9.98           \\
\texttt{accelerating}                 & \textbf{0.78}            & 7.67            & \textbf{1.52}            & 14.62          \\
\texttt{straight const @HS}           & \textbf{0.83}            & 6.92            & \textbf{1.64}            & 11.46          \\
\texttt{straight const @LS}           & \textbf{0.84}            & 8.39            & \textbf{1.64}            & 15.04          \\
\texttt{decelerating}                 & \textbf{0.79}            & 9.03            & \textbf{1.58}            & 14.82          \\ \midrule
\textbf{Average}                      & \textbf{0.81}            & 7.52            & \textbf{1.59}            & 13.75          \\ \bottomrule
\end{tabular}
\caption{\textbf{Fidelity of Estimated Vehicle Trajectories.} Comparison of ADE($\downarrow$) and FDE($\downarrow$) between our model and DROID-SLAM (DS), for estimated trajectories from video sequences. Our model shows consistently lower errors across all action categories. HS and LS represent high speed and low speed respectively. }

\label{tab:ade_fde_comparison}
\end{table}

\section{Action Controllability Exploration}
\label{sec:experiments}

This section explores action controllability for existing open-access world models: Vista and our model \textsc{Terra}.
Section \ref{sec:iec-eval} analyzes the instruction-execution consistency performance, and Section \ref{sec:ta_eval} evaluates estimated ego trajectories against the target trajectories provided during conditioning for assessing trajectory alignment.

We first summarize the process for generating action-conditioned scenes using Vista and \textsc{Terra}.
To effectively capture driving actions, a minimum duration of four seconds is considered essential. Vista is designed to generate sequences with a fixed input of three conditioning frames, producing 22 frames per round. Thus, we generate sequences over two rounds, resulting in a total of 44 frames (equivalent to 4.4 seconds of video).
\textsc{Terra} adopts the same setup as Vista for consistency.

Action conditioning for Vista is restricted to a single action input per generation round, limiting its flexibility. In contrast, \textsc{Terra} allows action conditioning on every frame, enabling trajectory inputs for each frame during the generation process. The frequency of instruction is determined by the specific constraints of each world model. 
As a result, each world model generates 2,286 videos using three conditioning frames and their corresponding trajectories from the \textsc{ACT-Bench} dataset.

\begin{figure}[t]
    \centering
    \includegraphics[width=1.0\linewidth]{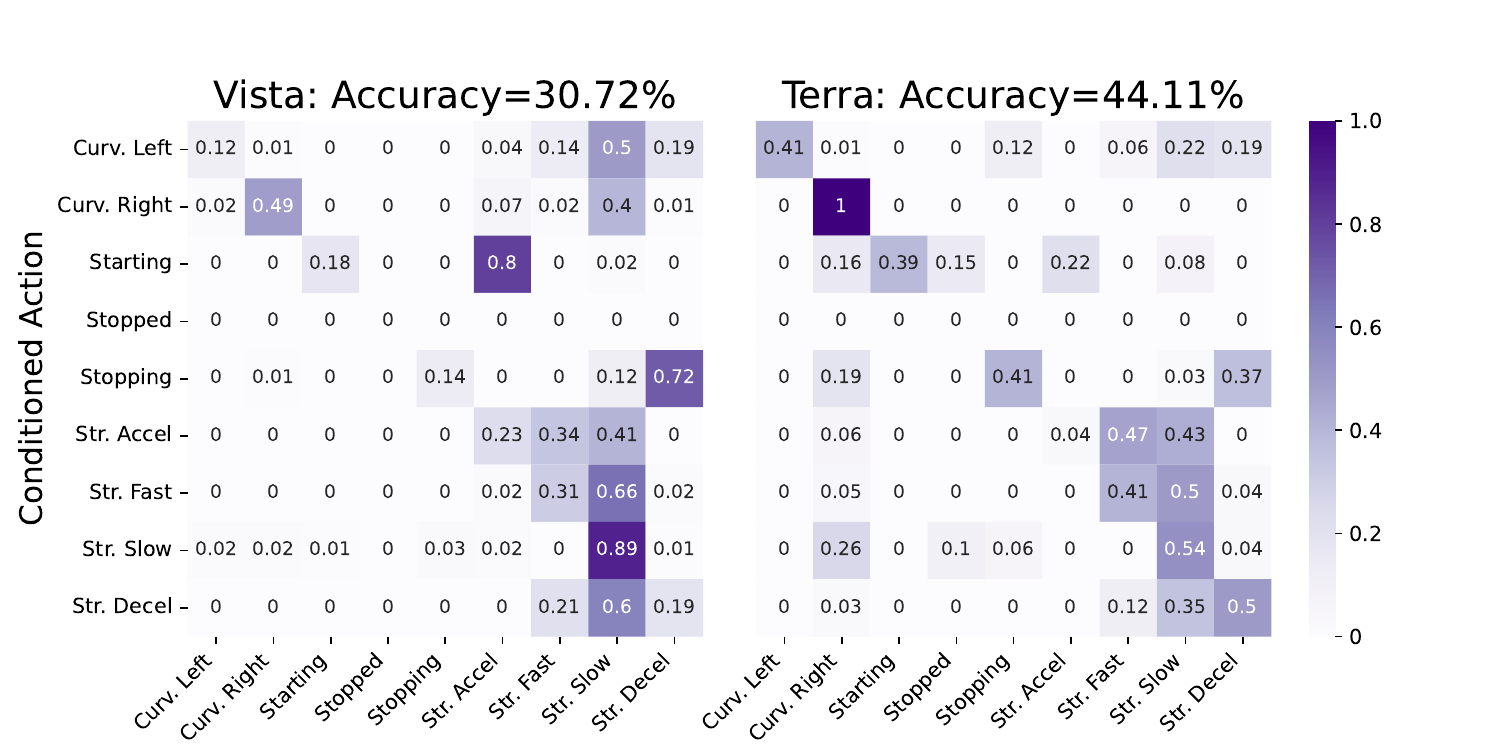}
    \caption{{\bf Instruction-Execution Consistency} for Vista and \textsc{Terra}. Vista struggles with curving actions, while \textsc{Terra} handles them more effectively, demonstrating a higher match rate.} 
    \label{fig:vista_conf_matrix}
\end{figure}

\begin{figure}[t]
    \centering
    \includegraphics[width=\linewidth]{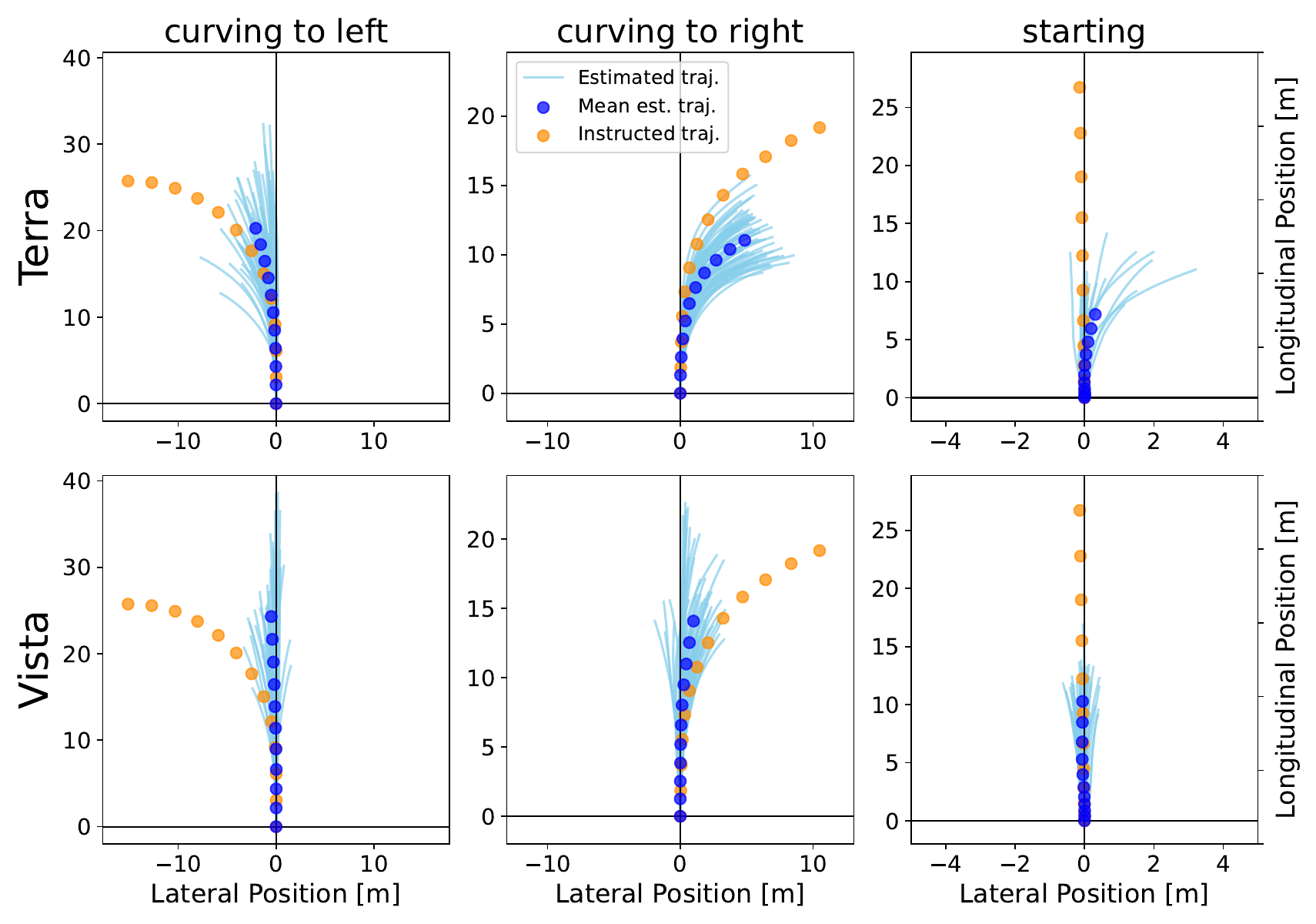}
    \caption{{\bf Trajectory Scatter Plot} for Vista and \textsc{Terra}. Vista struggles with curving actions, while \textsc{Terra} more effectively handles them, closely following the target trajectory's curvature.}
    \label{fig:estimated_trajectory_analysis}
\end{figure}

\begin{figure*}[t]
    \centering
    \includegraphics[width=1.0\linewidth]{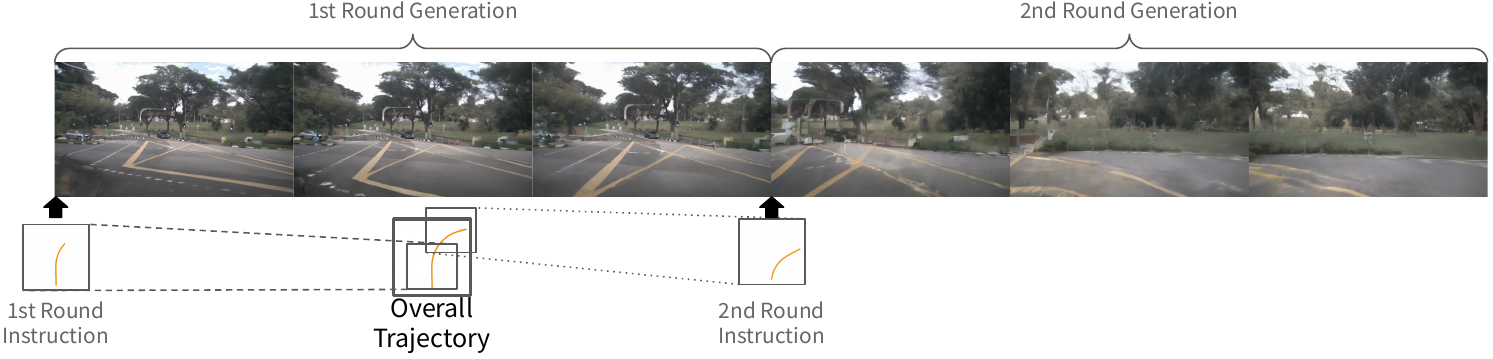}
    \caption{\textbf{An example of an abrupt, unnatural result in generation by Vista.} While the left three frames generated in the first round show nearly straight movement, the right three frames generated in the second round exhibit a significant shift to the right, resulting in a noticeably jerky motion when connected sequentially.}
    \label{fig:abrupt_motion}
    \includegraphics[width=1.0\linewidth]{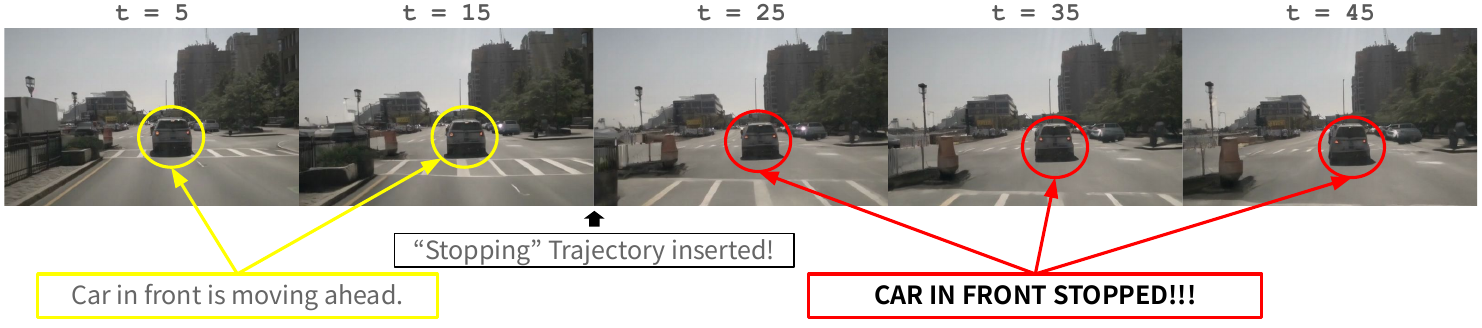}
    \caption{\textbf{An example of the \textit{Causal Misalignment} in world models.} Leading car stops in response to the "\texttt{stopping}" instruction given to the ego vehicle, but it is expected to continue its motion independently of the ego vehicle's actions.}
    \label{fig:causal_confusion}
    \includegraphics[width=1.0\linewidth]{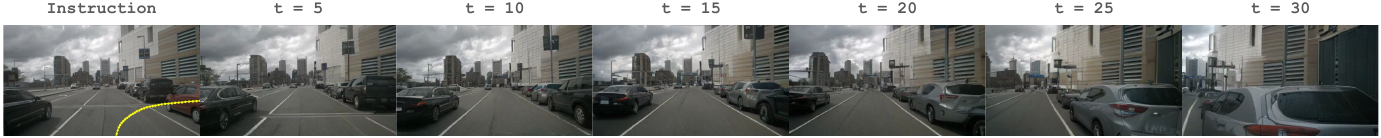}
    \caption{\textbf{An example of crash scene generated by \textsc{Terra}.} }
    \label{fig:crash_scene}
\end{figure*}

\subsection{Instruction-Execution Consistency Evaluation} \label{sec:iec-eval}

Figure \ref{fig:vista_conf_matrix} illustrates the ratios of estimated action occurrences compared to their ground-truth actions for Vista and \textsc{Terra}, respectively. 
Vista achieves a match rate of 30.72\%, while \textsc{Terra} achieves 44.11\%. 
Although \textsc{Terra} slightly outperforms Vista, both models show room for improvement in fidelity to conditioned action instructions.
As described in the experimental setup, \textsc{Terra} allows instruction injection at every frame, whereas Vista only permits it once per generation round. 
This difference in instruction frequency may have influenced these results.

\begin{table}[t]
\centering
\footnotesize
\begin{tabular}{l|cc|cc}
\toprule
\multirow{2}{*}{High-level Action Category} & \multicolumn{2}{c|}{ADE($\downarrow$)} & \multicolumn{2}{c}{FDE($\downarrow$)}    \\
                                   & Vista           & \textsc{Terra}  & Vista           & \textsc{Terra}  \\ \midrule
\texttt{curving to left}           & \textbf{3.59}   & 3.82            & 8.01            & \textbf{7.78}   \\
\texttt{curving to right}          & \textbf{3.73}   & 3.78            & 8.52            & \textbf{8.06}   \\
\texttt{starting}                  & \textbf{3.23}   & 4.52            & \textbf{10.32}  & 14.01           \\
\texttt{stopped}                   & N/A             & N/A             & N/A             & N/A             \\
\texttt{stopping}                  & 3.50            & \textbf{3.00}   & 3.97            & \textbf{3.91}   \\
\texttt{accelerating}              & 6.46            & \textbf{5.69}   & 14.98           & \textbf{14.17}  \\
\texttt{straight const @HS}        & 5.72            & \textbf{4.28}   & 11.28           & \textbf{9.25}   \\
\texttt{straight const @LS}        & 2.79            & \textbf{2.48}   & \textbf{5.34}   & 5.36            \\
\texttt{decelerating}              & 6.37            & \textbf{5.37}   & 11.56           & \textbf{10.53}  \\ \midrule
\textbf{Average}                   & 4.50            & \textbf{3.98}   & 8.66            & \textbf{8.21}   \\ \bottomrule
\end{tabular}
\caption{\textbf{Trajectory Alignment across Action Categories. } ADE($\downarrow$) and FDE($\downarrow$) are measured to evaluate how accurately each model generates motions that adheres to the conditioned trajectories across various high-level action categories. \textsc{Terra} results in better alignment to the conditioned trajectories against Vista.}
\label{tab:trajectory_alignment_evaluation}
\end{table}

\subsection{Trajectory Alignment Evaluation} \label{sec:ta_eval}

Table \ref{tab:trajectory_alignment_evaluation} shows ADE and FDE, which reveal how closely each model’s generated trajectories match the intended paths.
\textsc{Terra} outperforms Vista in both ADE and FDE across most high-level action classes, reflecting its superiority in responding to provided trajectory. 
These results confirm that \textsc{Terra} provides more accurate trajectory alignment capability.

Figure \ref{fig:estimated_trajectory_analysis} illustrates trajectory scatter plot results for these two world models.
From the perspective of curve accuracy, Vista fails to fully execute the curving action, resulting in a more straight-like trajectory. In contrast, Terra either closely follows the instructed trajectory or even produces a sharper curve, indicating that Terra is more proficient at generating curving actions than Vista. 
Although Terra shows slightly lower ADE for ``curving to left'' and ``curving to right,'' the visualized plot shows that \textsc{Terra} well performs instructed actions rather than Vista.
These observations suggest that \textsc{Terra} exhibits better action fidelity and controllability in response to given instructions. \textsc{Terra}’s frame-level conditioning supports more responsive curving and accurate directional adherence; however, it still faces challenges in maintaining consistent travel distance.

\subsection{Noteworthy Findings}

We find several remarkable findings in the analysis of Vista and \textsc{Terra} through our evaluation framework. 
The corresponding movies for Figure \ref{fig:abrupt_motion}, \ref{fig:causal_confusion}, and \ref{fig:crash_scene} are provided in supplemental materials.

First, we observe that Vista, which allows action conditioning only at each generation round, exhibit abrupt and unnatural motion changes at round transitions (Figure \ref{fig:abrupt_motion}).

Additionally, we discover cases where actions directed at the ego-vehicle appear to inadvertently influence other agents visible in the context, even though these actions are not intended to affect them. For example, when given instructions for the ego-vehicle to gradually decelerate and stop while following a car ahead, we observe that the car in front also came to an unexpected stop (Figure 
\ref{fig:causal_confusion}). Such behavior, termed \textit{Causal Misalignment}, which deviates from real-world dynamics, can pose a significant challenge when utilizing world models as simulators.

Finally, \textsc{Terra}'s high level controllability allows us to deliberately generate crash scenes (Figure \ref{fig:crash_scene}). This capability to produce difficult-to-collect scenarios highlights the potential of world models to provide  data for situations that are otherwise challenging to capture in real-world, thus enhancing their applicability in autonomous driving research.

\section{Conclusion}
\label{sec:conclusion}

To further advance action-controllable world models, we introduced \textsc{ACT-Bench}, an open-access evaluation framework for quantifying action fidelity. 
This framework includes an annotated dataset, a systematic evaluator model (\textsc{ACT-Estimator}), and a baseline world model (\textsc{Terra}). 
Leveraging this framework, we demonstrated that while the state-of-the-art model did not fully adhere to given instructions, \textsc{Terra} exhibited improved action fidelity. 
Additionally, we observed that \textsc{Terra} has the potential to generate diverse action-conditioned scenes, including cases of accidental collisions. 
We hope that our benchmark suite encourages further research and innovation in driving world models.

{
    \small
    \bibliographystyle{ieeenat_fullname}
    \bibliography{main}
}


\clearpage
\appendix
\section{Dataset Construction for ACT-Estimator}

This section details the procedure for constructing the dataset used to train our \textsc{ACT-Estimator}. The dataset is derived from the nuScenes dataset, specifically using sequences from the \texttt{CAM\_FRONT} sensor. Each sequence consists of 44 frames, corresponding to approximately four seconds of video, and includes the associated trajectory data computed from \texttt{ego\_pose} information.
To maximize dataset size while maintaining temporal coherence, overlapping windows with a stride of one frame are applied to slice the nuScenes frames into four-second segments. The trajectory data, representing the vehicle's position and orientation, is transformed into a local coordinate system centered on the initial frame of each segment for consistency.

High-level action labels are automatically assigned using a rule-based algorithm that categorizes trajectories into eleven predefined classes (see Table \ref{tab:dataset_details_for_act_estimator}). The algorithm uses thresholds for various features, including changes in waypoint interval distances, trajectory curvature, and the angle between the trajectory tangent and the y-axis (0° representing straight ahead). These thresholds are empirically calibrated to balance class distribution across the dataset, with detailed parameters provided in Table \ref{tab:trajectory_labels}.
This automated labeling process ensures accurate and consistent categorization of trajectories, making the dataset suitable for training and evaluating the \textsc{ACT-Estimator}.

\begin{table}[b]
    \centering
    \caption{Sample counts for each high-level action category in the dataset used to train the \textsc{ACT-Estimator}. The \textbf{Pred.} column indicates whether the category is included in the classification task, with excluded categories omitted due to class imbalance.}
    \label{tab:dataset_details_for_act_estimator}
    \small
    \begin{tabular}{l|c|c}
        \toprule
        \textbf{High-level Action Category} & \textbf{\#Samples} & \textbf{Pred.}   \\
        \midrule
        \texttt{curving to left}        & 7925   & \checkmark   \\
        \texttt{curving to right}       & 8264   & \checkmark   \\
        \texttt{shifting towards left}  & 285    &              \\
        \texttt{shifting towards right} & 353    &              \\
        \texttt{starting}               & 1952   & \checkmark   \\
        \texttt{stopped}                & 3958   & \checkmark   \\
        \texttt{stopping}               & 1809   & \checkmark   \\
        \texttt{accelerating}           & 1912   & \checkmark   \\
        \texttt{straight const @HS}     & 9055   & \checkmark   \\
        \texttt{straight const @LS}     & 8996   & \checkmark   \\
        \texttt{decelerating}           & 1903   & \checkmark   \\
        \midrule
        \textbf{Total} & \textbf{46412} \\
        \bottomrule
    \end{tabular}
\end{table}

\begin{table*}[h!]
\caption{\textbf{Conditions and Thresholds for Assigning High-Level Action Labels to Trajectories.} The labeling process applies each condition to the entire dataset sequentially, starting from the action with the fewest occurrences (e.g., \texttt{shifting\_towards\_left} and \texttt{shifting\_towards\_right}) and moving down the table to ensure that all matching trajectories are labeled.}

\label{tab:trajectory_labels}
\centering
\tiny
\begin{tabular}{|l|p{0.44\textwidth}|p{0.33\textwidth}|}
\hline
\textbf{Label}                    & \textbf{Condition}                                                                                                & \textbf{Thresholds}                                         \\ \hline
\texttt{shifting towards right}   & Lateral divergence (\texttt{lr\_div}) is positive and exceeds a threshold.                                        & \texttt{lr\_div > 1.3}                                      \\ 
                                  & The intermediate angle (\texttt{angle\_mid}) exceeds a threshold, indicating significant deviation.               & \texttt{angle\_mid > 4}                                     \\ 
                                  & The final angle (\texttt{angle\_last}) is below a threshold, indicating a return to the straight direction typical of lane changes. & \texttt{angle\_last < 2.3}                \\ \hline
\texttt{shifting towards left}    & Lateral divergence (\texttt{lr\_div}) is negative and exceeds a threshold (in the opposite direction).            & \texttt{lr\_div < -1.3}                                     \\ 
                                  & The intermediate angle (\texttt{angle\_mid}) exceeds a threshold, indicating significant deviation.               & \texttt{angle\_mid > 4}                                     \\ 
                                  & The final angle (\texttt{angle\_last}) is below a threshold, indicating a return to the straight direction typical of lane changes. & \texttt{angle\_last < 2.3}                \\ \hline
\texttt{curving to right}         & Trajectory length (\texttt{length}) exceeds a minimum threshold and is categorized into two ranges:               & \texttt{length > 3}                                         \\ 
                                  & - Short trajectories (3 to 10 points): Lateral divergence (\texttt{lr\_div}) meets a scaled threshold.             & \texttt{lr\_div >= 0.9 / 10 * length}                       \\ 
                                  & - Long trajectories (over 10 points): Lateral divergence meets a different scaled threshold.                       & \texttt{lr\_div >= 3.1 / 30 * length}                       \\ 
                                  & The trajectory is split into two halves, and circles are fitted to each half. The x-coordinates of the circle centers (\texttt{circle\_center\_x\_fh} and \texttt{circle\_center\_x\_lh}) are used to determine rightward curvature. & \texttt{circle\_center\_x\_fh > 0} and \texttt{circle\_center\_x\_lh > 0} \\ 
                                  & The starting interval (\texttt{closest\_interval}) exceeds a threshold, ensuring movement at the start.           & \texttt{closest\_interval > 0.005}                          \\ \hline
\texttt{curving to left}          & Trajectory length (\texttt{length}) exceeds a minimum threshold and is categorized into two ranges:               & \texttt{length > 3}                                         \\ 
                                  & - Short trajectories (3 to 10 points): Lateral divergence (\texttt{lr\_div}) meets a scaled threshold.             & \texttt{lr\_div <= -0.9 / 10 * length}                      \\ 
                                  & - Long trajectories (over 10 points): Lateral divergence meets a different scaled threshold.                       & \texttt{lr\_div <= -3.1 / 30 * length}                      \\ 
                                  & The trajectory is split into two halves, and circles are fitted to each half. The x-coordinates of the circle centers (\texttt{circle\_center\_x\_fh} and \texttt{circle\_center\_x\_lh}) are used to determine leftward curvature. & \texttt{circle\_center\_x\_fh < 0} and \texttt{circle\_center\_x\_lh < 0} \\ 
                                  & The starting interval (\texttt{closest\_interval}) exceeds a threshold, ensuring movement at the start.           & \texttt{closest\_interval > 0.005}                          \\ \hline

\texttt{starting}                 & Trajectory length (\texttt{length}) is within a specific short range, indicating the start of movement.           & \texttt{2 < length < 15}                                    \\ 
                                  & Closest interval (\texttt{closest\_interval}) is small, indicating a near-zero starting velocity.                 & \texttt{closest\_interval < 0.005}                          \\ 
                                  & Ratio of the first interval to the fourth (\texttt{interval\_1\_over\_4}) is small, suggesting gradual acceleration. & \texttt{interval\_1\_over\_4 < 0.05}                     \\ 
                                  & The change in interval distance (\texttt{interval\_delta}) is significant, indicating acceleration.               & \texttt{interval\_delta > 0.1}                              \\ \hline
\texttt{stopping}                 & Trajectory length (\texttt{length}) exceeds a minimum threshold.                                                  & \texttt{length > 3}                                         \\ 
                                  & Furthest interval (\texttt{furthest\_interval}) is small, indicating deceleration towards a stop.                 & \texttt{furthest\_interval < 0.03}                          \\ 
                                  & Ratio of the third interval to the fourth (\texttt{interval\_3\_over\_4}) is small, indicating deceleration.      & \texttt{interval\_3\_over\_4 < 0.08}                        \\ 
                                  & Closest interval (\texttt{closest\_interval}) is relatively large, suggesting the stop point.                     & \texttt{closest\_interval > 0.1}                            \\ 
                                  & The difference between closest and furthest intervals indicates significant deceleration.                         & \texttt{closest\_interval - furthest\_interval > 0.10}      \\ \hline
\texttt{stopped}                  & Trajectory length (\texttt{length}) is near zero, indicating no movement.                                         & \texttt{length < 0.01}                                      \\ \hline
\texttt{accelerating}             & Trajectory length (\texttt{length}) falls within specific ranges, with lateral divergence close to zero, indicating straight movement. & - Short trajectories (3-10 points): \texttt{abs(lr\_div) < 0.7 / 10 * length} \\ 
                                  &                                                                                                                   & - Long trajectories (10-44 points): \texttt{abs(lr\_div) < 2.5 / 30 * length} \\ 
                                  & Positive acceleration (\texttt{acceleration}) indicating increasing speed.                                        & - Length 3-20: \texttt{acceleration > 0.18}                 \\ 
                                  &                                                                                                                   & - Length 20-30: \texttt{acceleration > 0.3}                 \\ 
                                  &                                                                                                                   & - Length 30-35: \texttt{acceleration > 0.26}                \\ 
                                  & Closest interval (\texttt{closest\_interval}) is relatively large, suggesting ongoing movement.                   & \texttt{closest\_interval > 0.15}                           \\ \hline
\texttt{decelerating}             &  Trajectory length (\texttt{length}) falls within specific ranges, with lateral divergence close to zero, indicating straight movement. & - Length 5-15: \texttt{acceleration < -0.17}                \\ 
                                  &                                                                                                                   & - Length 15-25: \texttt{acceleration < -0.3}                \\ 
                                  &                                                                                                                   & - Length 25-40: \texttt{acceleration < -0.26}               \\ 
                                  &                                                                                                                   & - Length 40-55: \texttt{acceleration < -0.26}               \\ 
                                  & Interval delta (\texttt{interval\_delta}) is negative, indicating a reduction in distance between intervals.      & - Length 5-15: \texttt{interval\_delta < -0.2}              \\ 
                                  &                                                                                                                   & - Length 15-25: \texttt{interval\_delta < -0.23}            \\ 
                                  &                                                                                                                   & - Length 25-40: \texttt{interval\_delta < -0.21}            \\ 
                                  &                                                                                                                   & - Length 40-55: \texttt{interval\_delta < -0.4}             \\ 
                                  & Furthest interval (\texttt{furthest\_interval}) is sufficiently large to indicate that the trajectory is not coming to a complete stop, distinguishing it from stopping trajectories.  & \texttt{furthest\_interval > 0.15}      \\ \hline
\texttt{straight constant @LS}    & Trajectory length (\texttt{length}) is within a specific range, indicating low-speed movement (progressing at a constant low speed without stopping). & \texttt{3 < length < 25}    \\ 
                                  & Lateral divergence (\texttt{lr\_div}) is minimal, indicating straight movement.                                   & - Short trajectories (3-10 points): \texttt{abs(lr\_div) < 0.7 / 10 * length} \\ 
                                  &                                                                                                                   & - Long trajectories (10-44 points): \texttt{abs(lr\_div) < 2.5 / 30 * length} \\ 
                                  & The change in distances between consecutive trajectory points (\texttt{interval\_delta}) is minimal, indicating constant speed.  & \texttt{abs(interval\_delta) <= 0.5 / 40 * length}          \\ \hline
\texttt{straight constant @HS}    & Trajectory length (\texttt{length}) exceeds a minimum threshold, indicating high-speed movement (progressing at a constant high speed without stopping).  & \texttt{length > 28}     \\ 
                                  & Same lateral divergence conditions as \texttt{straight const @LS}.                                                & - Short trajectories (3-10 points): \texttt{abs(lr\_div) < 0.7 / 10 * length} \\ 
                                  &                                                                                                                   & - Long trajectories (10-44 points): \texttt{abs(lr\_div) < 2.5 / 30 * length} \\ 
                                  & The change in distances between consecutive trajectory points (\texttt{interval\_delta}) is minimal, indicating constant speed.  & \texttt{abs(interval\_delta) <= 0.5 / 40 * length}          \\ \hline

\end{tabular}
\end{table*}

\section{ACT-Estimator Design}

\begin{figure}[t]
    \centering
    \includegraphics[width=1.0\linewidth]{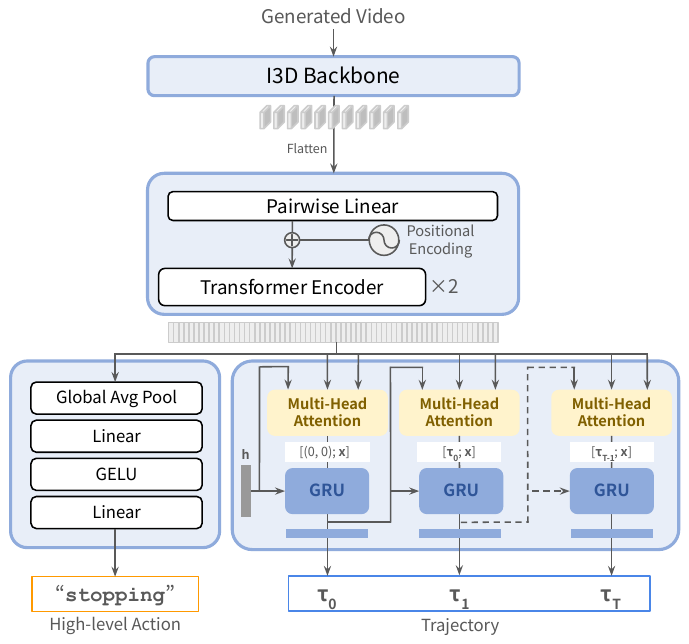}
    \caption{\textbf{Architecture of the \textsc{ACT-Estimator}.} Our motion estimator, \textsc{ACT-Estimator} model jointly performs high-level action classification and trajectory regression. It employs the I3D architecture \cite{carreira2017quo}, a well-established backbone, and incorporates a GRU-based decoder to autoregressively predict trajectories.}
    \label{fig:evaluator_model_architecture}
\end{figure}

The architecture of the \textsc{ACT-Estimator}, as shown in Figure~\ref{fig:evaluator_model_architecture}, is designed with simplicity and efficiency in mind. It combines an I3D backbone, a Transformer Encoder to refine spatio-temporal features, and task-specific heads to handle high-level action classification and trajectory regression tasks effectively. This lightweight design strikes a balance between performance and computational cost, enabling robust classification capability of high-level actions and trajectory estimation while remaining computationally efficient for inference.
The ``\textbf{Pred.}'' column in Table~\ref{tab:dataset_details_for_act_estimator} indicates the action classes that are included in the classification task of the \textsc{ACT-Estimator}. Notably, the classes \texttt{shifting towards left} and \texttt{shifting towards right} are excluded from the prediction targets due to their significantly smaller sample sizes, which each account for less than 1\% of the dataset. Including these classes could result in severe class imbalance, compromising overall classification performance. By excluding these categories, the model ensures reliable and consistent predictions across the remaining well-represented classes.

\section{Template Instruction Trajectory}
\begin{figure*}[t]
    \centering
    \includegraphics[width=1.0\linewidth]{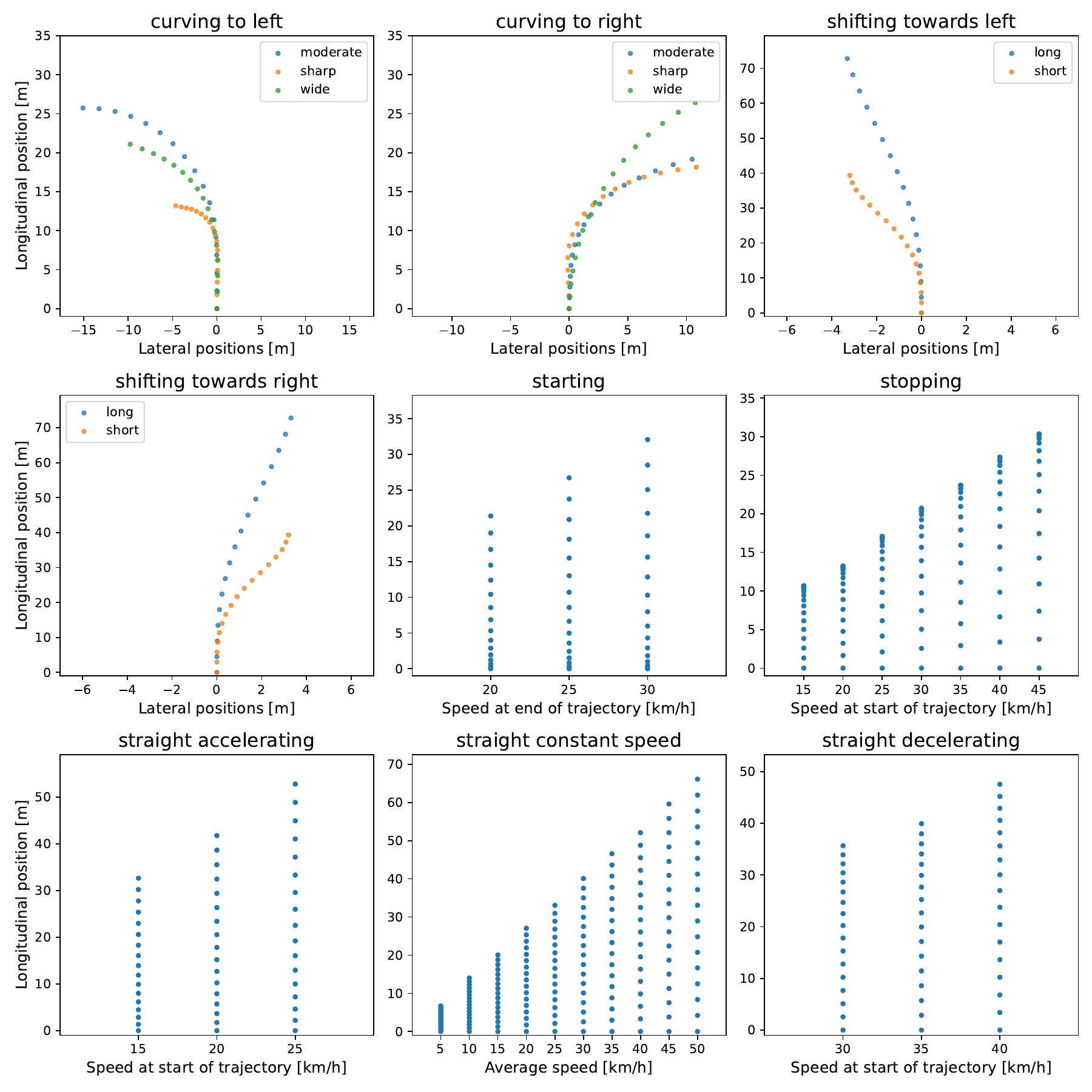}
    \caption{\textbf{Template Instruction Trajectories} used in the proposed \textsc{ACT-Bench}. The trajectories represent nine categories with 36 variations in total, showcasing diverse movement patterns. Each trajectory is manually selected and associated with a corresponding scene in \textsc{ACT-Bench} to ensure alignment with its intended instruction. These trajectories are carefully curated from the CoVLA dataset through a manual selection process to capture representative and meaningful motion behaviors.}
    \label{fig:instruction_trajectory}
\end{figure*}

Template instruction trajectories are essential for defining the ground truth vehicle movements in \textsc{ACT-Bench} and evaluating the fidelity of generated driving scenes. These trajectories act as ground truth references, paired with context videos extracted from the nuScenes dataset. Figure~\ref{fig:instruction_trajectory} illustrates all the instruction trajectories used in \textsc{ACT-Bench}. To ensure consistency, the initial speed of each trajectory is matched with the starting speed of the vehicle in the context video.

To achieve comprehensive evaluation, we defined nine categories of instruction trajectories, such as curving, lateral shifting, starting, stopping, accelerating, and maintaining constant speed. Each category includes multiple variations based on curvature, speed, or displacement, resulting in 36 distinct trajectories. This diversity ensures that \textsc{ACT-Bench} effectively assesses world models' ability to generate realistic and instruction-adherent driving scenarios.

\section{\textsc{Terra} World Model Design}
\label{sec:world-model}

As illustrated in Figure \ref{fig:architecture_overview}, \textsc{Terra} is an autoregressive Transformer-based World Model that takes a sequence of discretized image tokens and a vector sequence of trajectory instructions as input, predicting the sequence of image tokens at future time steps. An Image Tokenizer is employed to convert a sequence of image frames into a sequence of discrete tokens, while a Frame-wise Decoder is used to transform the sequence of discrete tokens back into image frames. These components correspond to the Encoder and Decoder of an Autoencoder, respectively. The vector sequence of trajectory instructions is processed through an Action Embedder for input representation. Additionally, \textsc{Terra} incorporates a post-hoc Video Refiner to enhance temporal consistency and resolution of the videos predicted by the Frame-wise Decoder. The following section, \ref{subsec:world_modeling_through_token_prediction}, provides a detailed description of the core components of the world model, while Section \ref{subsec:video_refiner} focuses specifically on the design and functionality of the Video Refiner.

\subsection{World Modeling through Token Prediction}
\label{subsec:world_modeling_through_token_prediction}

\paragraph{Image Tokenization.}

Given that our world model is constructed using an autoregressive Transformer, which works well with discrete token representations, we opt to represent the latent codes as sequences of discrete tokens. Formally, we employ CNN based encoder network $E_\theta$ to tokenize a sequence of frames $\mathbf{X} = (\mathbf{x}_1, \dots, \mathbf{x}_T)\in \mathbb{R}^{T\times H \times W \times 3}$ into discrete latent codes $\mathbf{C} = (\mathbf{c}_1, \dots,\mathbf{c}_T)\in\{1, 2, \dots, K\}^{T\times H^\prime \times W^\prime}$. Here, $H$ denotes the height of the image, $W$ the width of the image, and $K$ the codebook (vocabulary) size. Let $H^\prime$ and $W^\prime$ represent the downscaled dimensions, defined as $H^\prime = H / D$ and $W^\prime = W/D$, where $D$ is the downscaling factor. Each $\mathbf{c}_t$ is subsequently flattened into a one-dimensional sequence of discrete tokens in raster-scan order before being input into the autoregressive Transformer. 

\paragraph{Action-conditioning with Sequence of Trajectories.}

Since \textsc{Terra} aims to be utilized as a simulator for autonomous driving, it is designed to accept future vehicle trajectories as input, a format commonly adopted as the output by many autonomous driving planning algorithms\cite{hu2023planning, jiang2023vad, chen2024vadv2, weng2024drive}. A trajectory is provided at each time step $t$, represented as a sequence $\mathbf{a}_t=(\mathbf{a}^t_1, \dots, \mathbf{a}^t_L)$, where each point $\mathbf{a}^t_l = (x_l, y_l, t_l)$ indicates the vehicle's position in a vehicle-centered coordinate system $t_l$ seconds into the future, with the vehicle's position at time step $t$ as the origin. 

\paragraph{Interleaved Inputs.}

During training phase, it is assumed that trajectories $\textbf{a}_1,\dots,\textbf{a}_T$ are available at each time step corresponding to the discrete codes $\textbf{c}_1, \dots, \textbf{c}_T$ of the $T$ frames. In cases where corresponding trajectory data is not available, a special trajectory $\textbf{a}^* = (\textbf{a}^*_1, \dots, \textbf{a}^*_L)$ representing an empty trajectory is used for all time steps $t$. Latent codes and trajectories are then arranged in an interleaved format as $(\mathbf{c}_1, \mathbf{a}_1, \mathbf{c}_2, \mathbf{a}_2, \dots, \mathbf{c}_T,\mathbf{a}_T)$. On the other hand, in the inference phase, we autoregressively predict the discrete codes $\hat{\mathbf{c}}_{T^\prime + 1}, \dots, \hat{\mathbf{c}}_T$ of the subsequent frames using the discrete codes $\mathbf{c}_1, \dots, \mathbf{c}_{T^\prime}$ of the frame sequence provided as context, along with the trajectories $\mathbf{a}_1, \dots, \mathbf{a}_T$ for $T (> T^\prime)$ time steps. Initially, we provide $(\mathbf{c}_1, \mathbf{a}_1,\dots,\mathbf{c}_{T^\prime}, \mathbf{a}_{T^\prime})$ and predict the discrete code sequence $\hat{\mathbf{c}}_{T^\prime + 1}$, generating one token at a time. After predicting $N = H^\prime \times W^\prime$ tokens, we insert $\mathbf{a}_{T^\prime + 1}$ afterward, reformulating the sequence as $(\mathbf{c}_1,\mathbf{a}_1,\dots, \mathbf{c}_{T^\prime}, \mathbf{a}_{T^\prime}, \hat{\mathbf{c}}_{T^\prime + 1}, \mathbf{a}_{T^\prime + 1})$, thereby enabling the prediction of discrete tokens for the next frame.

Before being fed into the autoregressive Transformer, the data are first transformed into embeddings of $d$ dimensions. The token sequences representing image frames, given their discrete nature, are embedded through a learnable lookup table. In contrast, each trajectory $\mathbf{a}_t$ consists of $L$ three dimensional vectors representing future positions and timestamps, and is therefore converted into embeddings via a linear layer.

\paragraph{Learnable Positional Embedding.}

We apply learnable positional encodings decomposed into temporal and spatial components. The temporal positional encoding provides $d$-dimensional embeddings that assign unique values at each time step $t$ for the image frames. In contrast, the spatial positional encoding assigns unique values to each of the $N + L$ tokens within the same time steps.

\paragraph{Training Objective.}

The autoregressive Transformer is trained on next token prediction task. In this process, the loss is computed only for the token sequences representing image frames, while tokens representing trajectories are excluded from loss calculation. The loss function is formalized as follows.

\begin{equation}
    \mathcal{L}_{\mathrm{world~model}} = -\sum^T_{t=1} \sum^{N}_{n=1} \log p(c_{t,n} | \mathbf{c}_{<t},c_{t,m < n},\mathbf{a}_{<t})
\end{equation}

\subsection{Video Refiner}
\label{subsec:video_refiner}

When employing the world model as a simulator for camera-based autonomous driving systems, it becomes necessary to decode predicted future states, represented as discrete token sequences, back into video sequences. A straightforward approach to achieve this is to utilize the decoder from the image tokenizer, which is typically adopt an Autoencoder architecture, to decode each frame individually. However, with this approach, the resulting video may exhibit low temporal consistency, even if the quality of individual frames is high. Furthermore, when downscaling is applied to the images to reduce the sequence length input to the autoregressive model, the decoded images are also downscaled, which is suboptimal for use as a neural simulator. To address these issues and improve both image resolution and temporal consistency, we employ a latent diffusion model (LDM)~\cite{podell2023sdxl,rombach2022high} based Video Refiner. Specifically, the Video Refiner is constructed by fine-tuning the pre-trained model of Stable Video Diffusion (SVD)~\cite{blattmann2023stable}, an image-to-video model. In SVD, the conditioning image is first transformed into a latent representation $\mathbf{z}\in\mathbb{R}^{C_r\times H_r\times W_r}$, which is then concatenated along the channel axis with each frame of the noise $\mathbf{n}\in\mathbb{R}^{T_r \times C_r \times H_r \times W_r}$, resulting in a combined latent representation $\mathbf{n}^\prime \in \mathbb{R}^{T_r \times 2C_r \times H_r \times W_r}$. By iteratively denoising $\mathbf{n}^\prime$ using the U-net model $D_\theta$, a video is generated with reference to conditioning image. On the other hand, we first decode images using the Autoencoder's decoder, then upscale it to the desired resolution, and use this sequence of frames as conditioning. Conditioning images are transformed into latent representations $\mathbf{z^\prime} \in\mathbb{R}^{T_r\times C_r \times H_r \times W_r}$ by the VAE encoder of SVD, which are then concatenated to the noise $\mathbf{n}.$ The training and inference process follows the same flow as SVD.

\section{Implementation Details of \textsc{Terra}}

\subsection{Hyper-parameter Settings}

We set the size of the input images before passing them into the Image Tokenizer to $H = 288$ and $W = 512$. During training, we process 25 frames at a time ($T = 25$). Since we handle videos at a frame rate of 10 Hz, this corresponds to 2.5 seconds of video. For algorithms that convert image(s) into sequence(s) of discrete tokens, VQ-VAE~\cite{van2017neural} is widely known; however, we employ a more expressive approach using Lookup-Free Quantization~\cite{yulanguage}. Specifically, we utilize the pre-trained weights\footnote{\url{https://huggingface.co/TencentARC/Open-MAGVIT2/blob/2f7982b9d1d4c540645a5fb2c39e5892ebea15b7/imagenet\_256\_B.ckpt}} of Open-MAGVIT2~\cite{luo2024open} as our tokenizer. The Image Tokenizer we utilize is configured with a codebook size of $K = 262,144$ and the downscaling parameter of $D = 16$. As a result, the number of discrete tokens used to represent a single image is $N = 288 / 16 \times 512 / 16 = 576$. The length of the vector sequence representing actions is $L = 6$, resulting in a sequence length during training of $(N + L) \times T = (576 + 6) \times 25 = 14,550$. The dimensionality of the embedding input to the Transformer is set to $d = 2048$. As a special trajectory $\textbf{a}^*$ used for padding, we employ a matrix where all elements are set to $-1.0$:
$$
\left[
\begin{array}{ccc}
    -1. & -1. & -1. \\
    -1. & -1. & -1. \\
    -1. & -1. & -1. \\
    -1. & -1. & -1. \\
    -1. & -1. & -1. \\
    -1. & -1. & -1.
\end{array}
\right]
$$

\noindent The values of $t_l$ vary depending on the training dataset, as shown below:

\begin{equation*}
t_l = \begin{cases}
    0.45 + 0.5 \times (l - 1), \quad l = 1, 2, \ldots, 6. & \text{(CoVLA)} \\
    0.5 \times l, \quad l = 1, 2, \ldots, 6. & \text{(nuScenes)}
\end{cases}
\end{equation*}

\noindent In the Video Refiner, the images are first upscaled from $288 \times 512$ to $384 \times 640$. Subsequently, the latent variables compressed to $H_r = 48$ and $W_r = 80$ using the pre-trained Autoencoder from SVD are utilized. The settings for $T_r$ and $C_r$ are kept consistent with those of SVD, using $T_r = 25$ and $C_r = 4$.

\subsection{Training Procedure}

We conduct the training of the world model and the Video Refiner separately. As a preparation step for training both models, videos from OpenDV-YouTube, nuScenes, and the CoVLA dataset are converted into sequences of image frames at 10 Hz. Each image frame is subsequently transformed into a sequence of tokens using the pre-trained Image Tokenizer. For CoVLA dataset, since trajectory data in the vehicle-centric coordinate systems is available for each time step up to 2.95 seconds ahead, trajectory instruction data is created by sampling six $(x, y)$ coordinates. Similarly, for the nuScenes dataset, trajectory data in the vehicle-centric coordinate system is generated based on the \texttt{ego\_pose} up to 3 seconds ahead, from which six $(x, y)$ coordinates is sampled to create the trajectory instruction data. The data is segmented into non-overlapping chunks of 25 frames each and stored. Ultimately, the OpenDV-YouTube dataset is divided into 1.67 million chunks, the nuScenes dataset into 25,000 chunks, and the CoVLA dataset into 0.23 million chunks.

We employ a Transformer based on the Llama~\cite{touvron2023llama} architecture as the world model, which is trained from a randomly initialized state. The training is conducted over 40k steps using 56 H100 80GB GPUs, with a per-GPU batch size of 1. Gradient accumulation steps is set to 4. The world model is optimized using AdamW~\cite{loshchilov2017decoupled} optimizer in combination with a Cosine Decay learning rate schedule. The detailed parameter settings for the world model training are provided in the Table \ref{table:hyperparameters}.

\begin{table}[b]
    \centering
    \small
    \caption{Hyper-parameter settings for the world model training}
    \begin{tabular}{|l|l|}
        \hline
        \textbf{Model Parameters}            & \textbf{Value}                \\ \hline
        vocab\_size                          & 262145                        \\ 
        hidden\_size                         & 2048                          \\ 
        intermediate\_size                   & 5632                          \\ 
        num\_hidden\_layers                  & 22                            \\ 
        num\_attention\_heads                & 32                            \\ 
        num\_key\_value\_heads               & 4                             \\ 
        max\_position\_embeddings            & 14550                         \\ 
        activation\_function                 & ``relu"                        \\ 
        attention\_dropout                   & 0.0                           \\ 
        attn\_implementation                 & ``flash\_attention\_2"         \\ 
        pad\_token\_id                       & 262144                        \\ 
        bos\_token\_id                       & 262144                        \\ 
        eos\_token\_id                       & 262144                        \\ \hline
        \textbf{Optimizer Parameters}        & \textbf{Value}                \\ \hline
        type                                 & AdamW                         \\ 
        learning\_rate                       & 1.0e-4                        \\ 
        betas                                & (0.9, 0.999)                  \\ 
        weight\_decay                        & 0.0                           \\ 
        eps                                  & 1e-8                          \\ \hline
        \textbf{Learning Rate Scheduler Parameters} & \textbf{Value}          \\ \hline
        type                                 & cosine                        \\ 
        num\_warmup\_steps                   & 0                             \\ 
        num\_training\_steps                 & 172440                        \\ \hline
    \end{tabular}
    \label{table:hyperparameters}
\end{table}

The training of the Video Refiner is based on the first-stage training setup of Vista~\cite{gao2024vista}. In Vista, a dynamic prior is provided for the first three frames, and the initial frame is used as a conditioning frame by concatenating it with a noise tensor $\mathbf{n}$ along the channel axis. However, in our case, the goal is to refine the coarse predictions mode by the Frame-wise Decoder. Therefore, we do not include a dynamic prior. Instead, we concatenate the latent variables of the coarse predictions for each frame, as predicted by the Frame-wise Decoder, with the noise tensor $\mathbf{n}$ along the channel axis. The training is conducted over 800k steps on 8 H100 80GB GPUs with a per-GPU batch size of 1.
\section{Video Generation Settings}

For video generation with Vista, we refer to the \texttt{sample.py}\footnote{\url{https://github.com/OpenDriveLab/Vista/blob/main/sample.py}} script in Vista and use the parameter settings listed in the Table \ref{tab:vista_hyperparams}. However, in the case of multi-round generation with Vista, the same instructions are repeatedly used for each generation round. In our dataset, corrected target trajectories are provided for each future frame, representing the position and orientation at each timestep if the vehicle were to move faithfully along the target trajectory. Therefore, we modify the process of multi-round generation to use the trajectory assigned to the frame at the start of each round as illustrated in Figure \ref{fig:abrupt_motion}.

\begin{table}[h]
    \centering
    \caption{Hyper-parameter Settings for Video Generation with Vista}
    \small
    \begin{tabular}{|l|l|}
        \hline
        \textbf{Parameter} & \textbf{Value} \\ \hline
        action             & ``traj"         \\ 
        n\_rounds          & 2              \\ 
        n\_frames          & 25             \\ 
        n\_conds           & 1              \\ 
        seed               & 23             \\ 
        height             & 576            \\ 
        width              & 1024           \\ 
        cfg\_scale         & 2.5            \\ 
        cond\_aug          & 0.0            \\ 
        n\_steps           & 50             \\ \hline
    \end{tabular}
    \label{tab:vista_hyperparams}
\end{table}

In video generation with Terra, trajectory instructions are incorporated by appending the trajectory instruction $\mathbf{a}_t$ corresponding to each frame to the sequence of image tokens $\hat{\mathbf{c}}_t$ generated for that timestep. This approach is repeated for every frame during the generation process. To accelerate inference, video generation is performed using vLLM~\cite{kwon2023efficient}. We conduct generation with the generation parameter settings $\texttt{temperature}=0.9$, $\texttt{top\_p} = 1.0$ and $\texttt{top\_k} = -1$.
\section{Visualization}

\begin{figure*}[t!]
    \centering
    \includegraphics[width=1.0\linewidth]{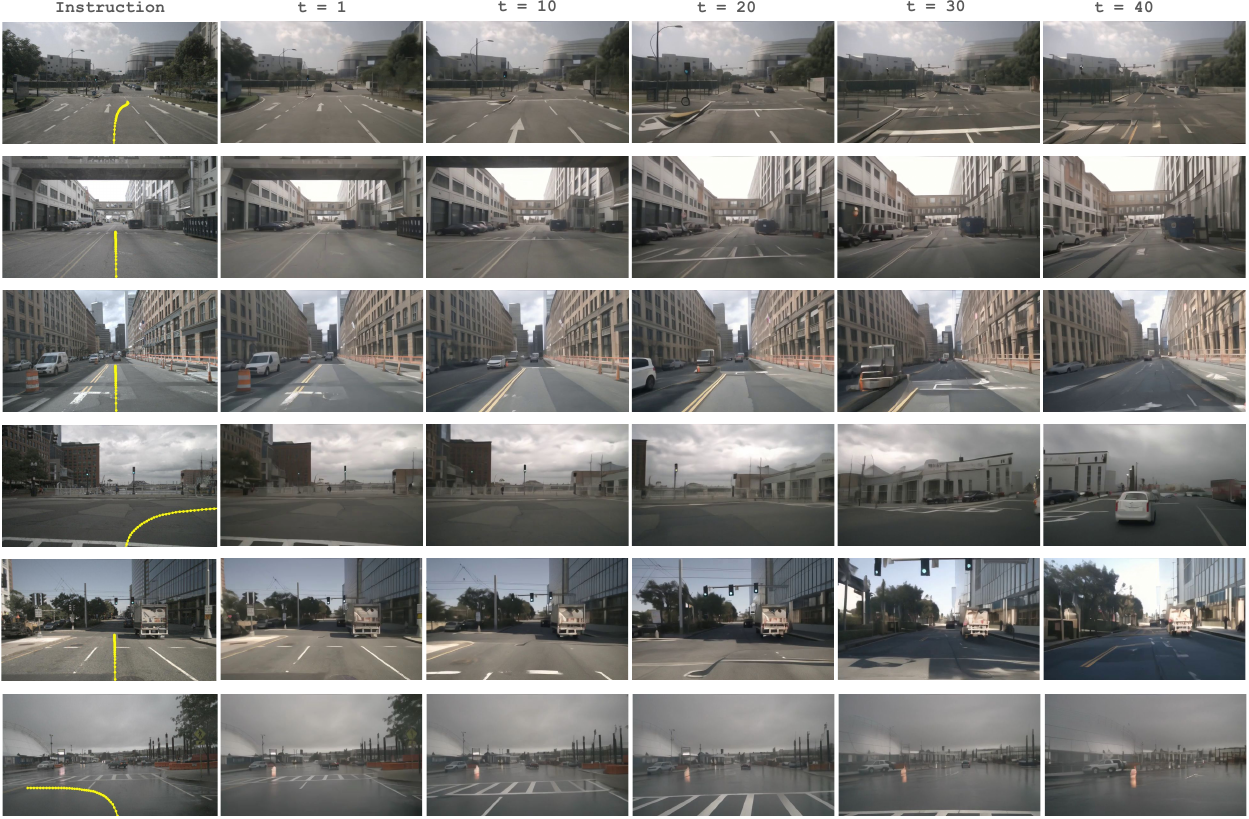}
    \caption{\textbf{Generation Capability of \textsc{Terra}.} Examples of video generation results by \textsc{Terra}, showing its ability to generate realistic driving scenes that adhere to specific instructions. The leftmost column visualizes the provided instruction trajectory, and the subsequent columns depict generated frames corresponding to the instruction at various time steps.}
    \label{fig:terra_video_examples}
\end{figure*}

Figure \ref{fig:terra_video_examples} visualizes videos generated by Terra. While the movements do not exactly follow the instructed trajectory, they demonstrate a reasonable level of adherence to the given instructions.

\begin{figure*}[t!]
    \centering
    \includegraphics[width=1.0\linewidth]{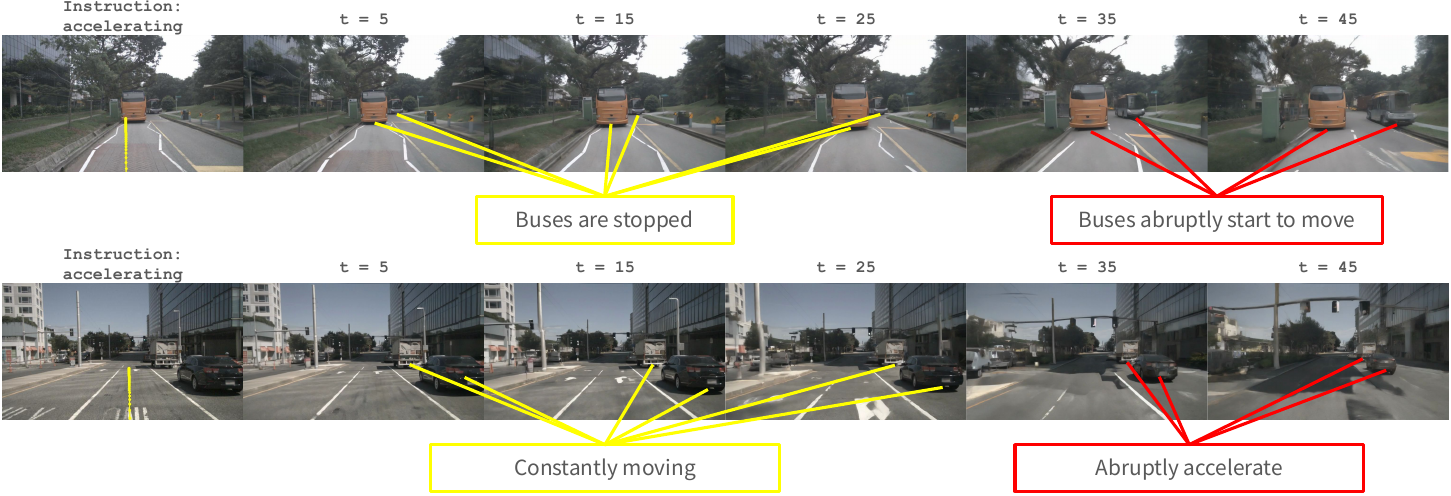}
    \caption{\textbf{\textit{Causal Misalignment} examples in generated videos of Vista.} Example in the first row illustrates a case where the preceding and oncoming vehicle starts moving unnaturally as the ego vehicle approaches. In contrast, example in the second row depicts a scenario where a parallel vehicle accelerates unnaturally. In Vista, instructions are inserted at the transitions between rounds, making these transitions particularly prone to noticeable changes.}
    \label{fig:vista_causal_misalignment}
    \vspace{15mm}
    \centering
    \includegraphics[width=1.0\linewidth]{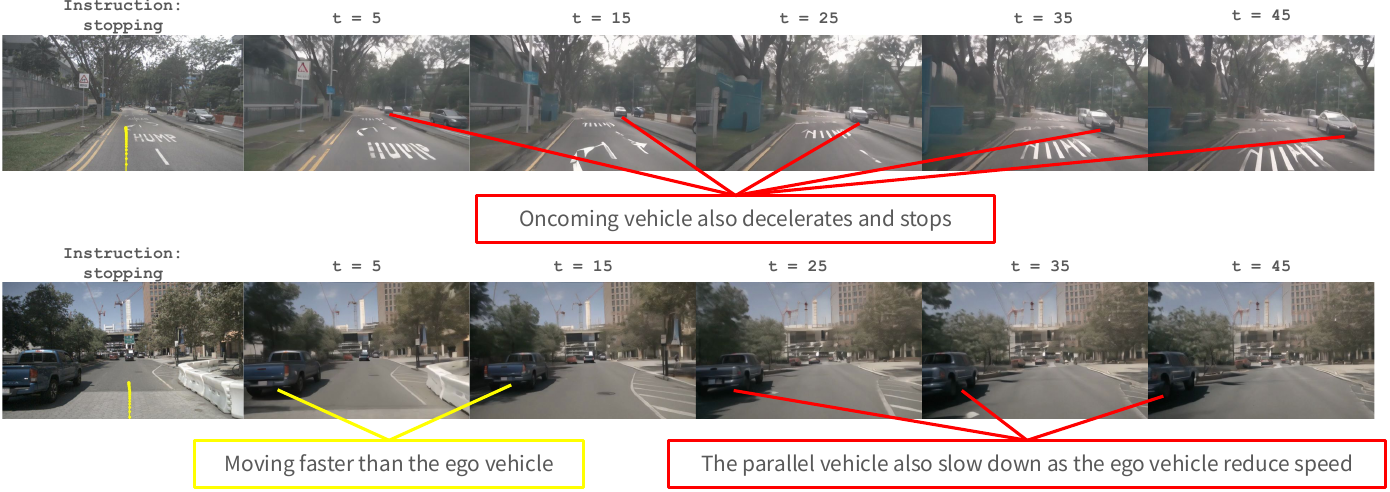}
    \caption{\textbf{\textit{Causal Misalignment} examples in generated videos of \textsc{Terra}.} The example in the first row demonstrates a scenario where the oncoming vehicle gradually decelerates and comes to a stop in response to the deceleration of the ego vehicle. In the second-row example, the parallel vehicle, initially moving faster than the ego vehicle, similarly decelerates and eventually stops as the ego vehicle reduces its speed.}
    \label{fig:terra_causal_misalignment}
\end{figure*}

In Figure \ref{fig:vista_causal_misalignment}, the first row illustrates a case where the preceding and oncoming vehicles begin moving unnaturally as the ego vehicle approaches. In contrast, the example in the second row depicts a scenario where a parallel vehicle accelerates unnaturally. In Vista, instructions are inserted at the transitions between rounds, making these transitions particularly prone to noticeable irregularities.

Figure \ref{fig:terra_causal_misalignment} demonstrates an example where the oncoming vehicle gradually decelerates and comes to a stop in response to the ego vehicle's deceleration in the first-row example. In the second-row example, the parallel vehicle, initially moving faster than the ego vehicle, similarly decelerates and eventually stops as the ego vehicle reduces its speed.

\end{document}